\documentclass[runningheads]{llncs}
\usepackage{graphicx}
\usepackage{subfigure}
\usepackage{makecell}
\usepackage{amsmath,amssymb,amsfonts}
\usepackage[colorlinks,
urlcolor=blue,
linkcolor=blue,      
anchorcolor=blue,  
citecolor=blue,     
]{hyperref}
\usepackage{url}
\usepackage[T1]{fontenc}

\begin{document}
\title{A Correction-Based Dynamic Enhancement Framework towards Underwater Detection}
\titlerunning{A Utility-Oriented Underwater Enhancement Framework}
\author{Yanling Qiu\and
Qianxue Feng \and
Boqin Cai\and
Hongan Wei\and
Weiling Chen}

\authorrunning{Y. Qiu et al.}

\institute{Fujian Key Lab for Intelligent Processing and Wireless Transmission of Media Information, Fuzhou University, Fuzhou 350108, China\\
\email{\{201120096, 211120125, 221120099, weihongan, weiling.chen\}@fzu.edu.cn}}
\maketitle              

\begin{abstract}
To assist underwater object detection for better performance, image enhancement technology is often used as a pre-processing step. However, most of the existing enhancement methods tend to pursue the visual quality of an image, instead of providing effective help for detection tasks. In fact, image enhancement algorithms should be optimized with the goal of utility improvement. In this paper, to adapt to the underwater detection tasks, we proposed a lightweight dynamic enhancement algorithm using a contribution dictionary to guide low-level corrections. Dynamic solutions are designed to capture differences in detection preferences. In addition, it can also balance the inconsistency between the contribution of correction operations and their time complexity. Experimental results in real underwater object detection tasks show the superiority of our proposed method in both generalization and real-time performance. 


\end{abstract}
\section{Introduction}
\begin{figure}[t]
	\begin{minipage}{0.5\linewidth}
		\centering 
		\subfigure[{Low light}]{
			\label{Fig. 1. sub. 1}
			\includegraphics[width=2.5cm,height=2.5cm]{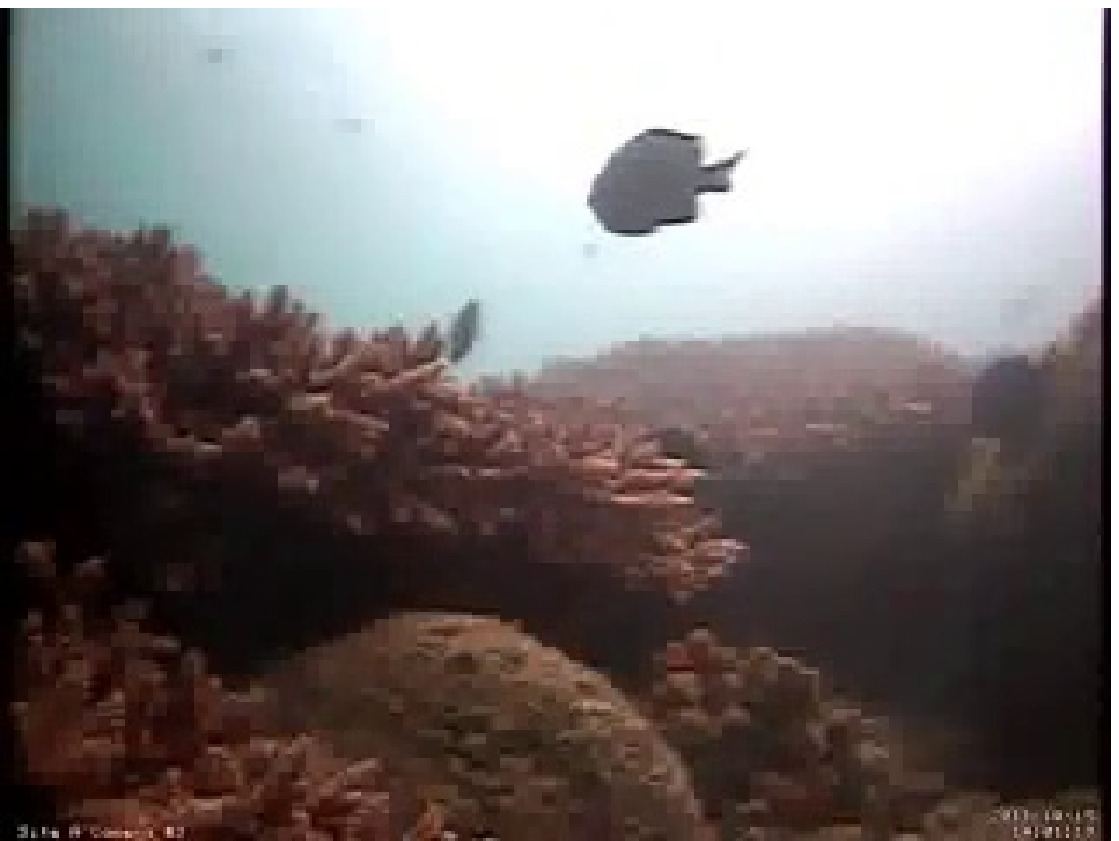}}
		\hspace{-0.3cm}
		\subfigure[{Blue tone}]{
			\label{Fig. 1. sub. 2}
			\includegraphics[width=2.5cm,height=2.5cm]{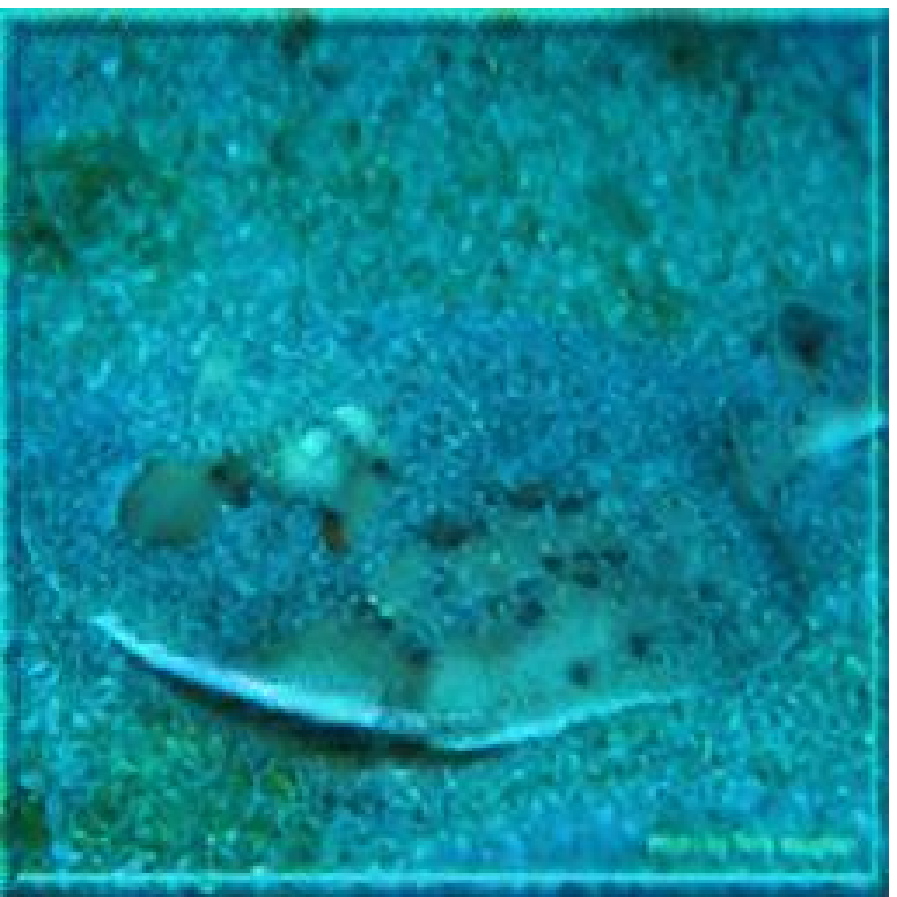}}
		\qquad
		\subfigure[{Ocean snow}]{
			\label{Fig. 1. sub. 3}
			\includegraphics[width=2.5cm,height=2.5cm]{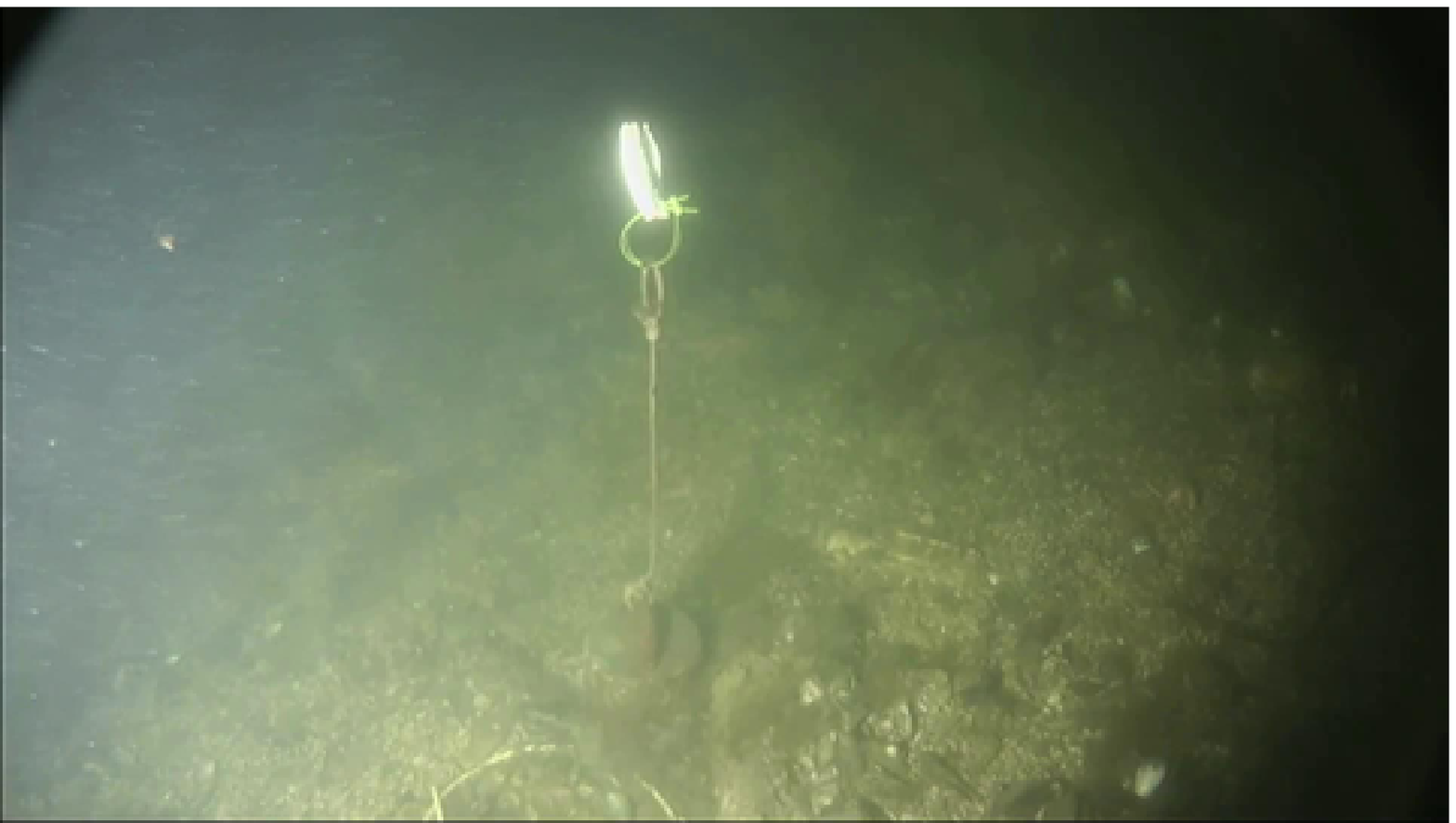}}
		\hspace{-0.3cm}
		\subfigure[{Motion blur}]{
			\label{Fig. 1. sub. 4}
			\includegraphics[width=2.5cm,height=2.5cm]{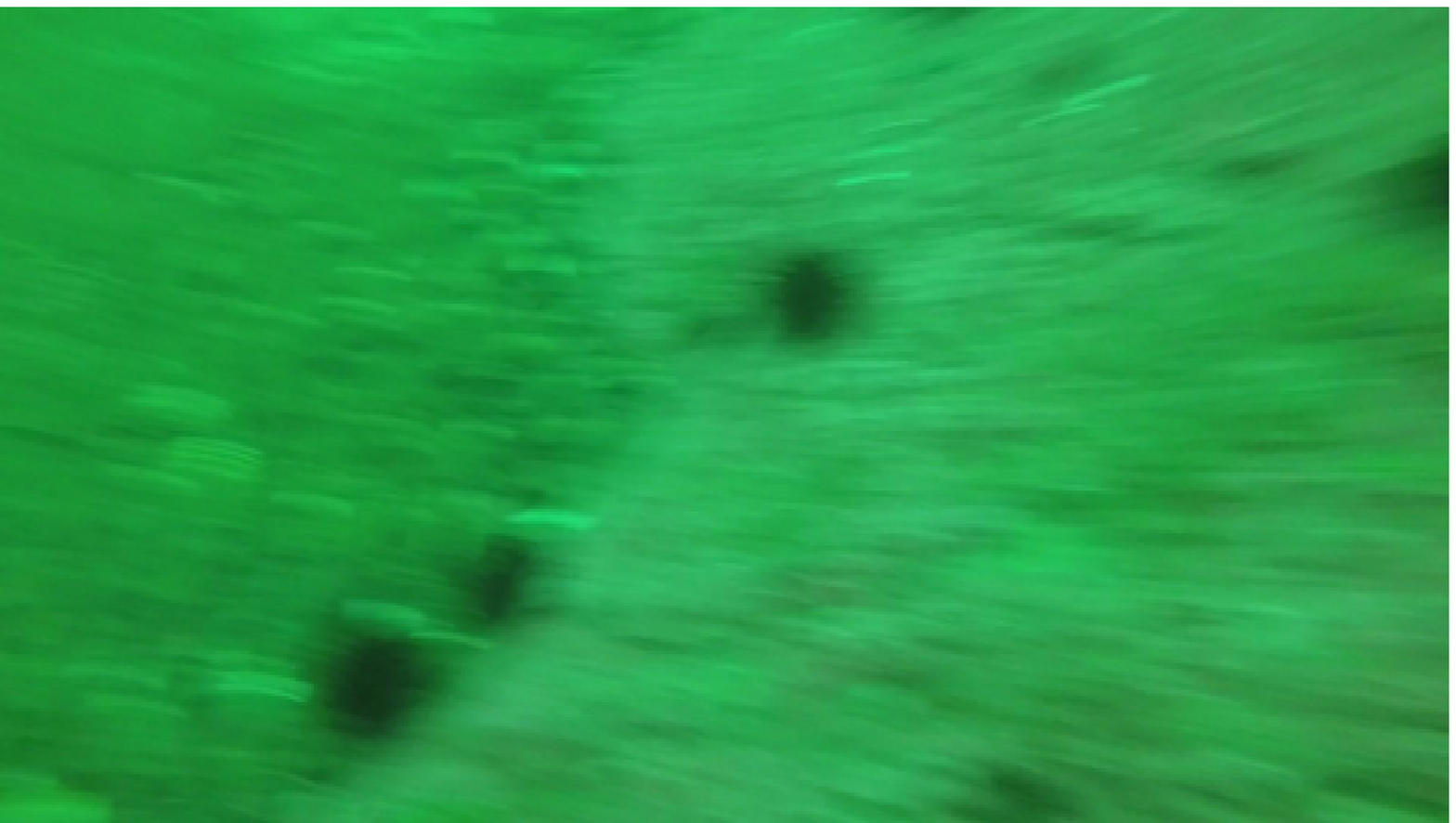}}
		\caption{Degraded underwater images.} 
		\label{Fig. 1}
	\end{minipage}
	\hspace{-0.6cm}
	\begin{minipage}{0.5\linewidth}
		\centering 
		\subfigure[Image visual quality]{
			\label{Fig. 2. sub. 1}
			\includegraphics[width=5.5cm,height=2.6cm]{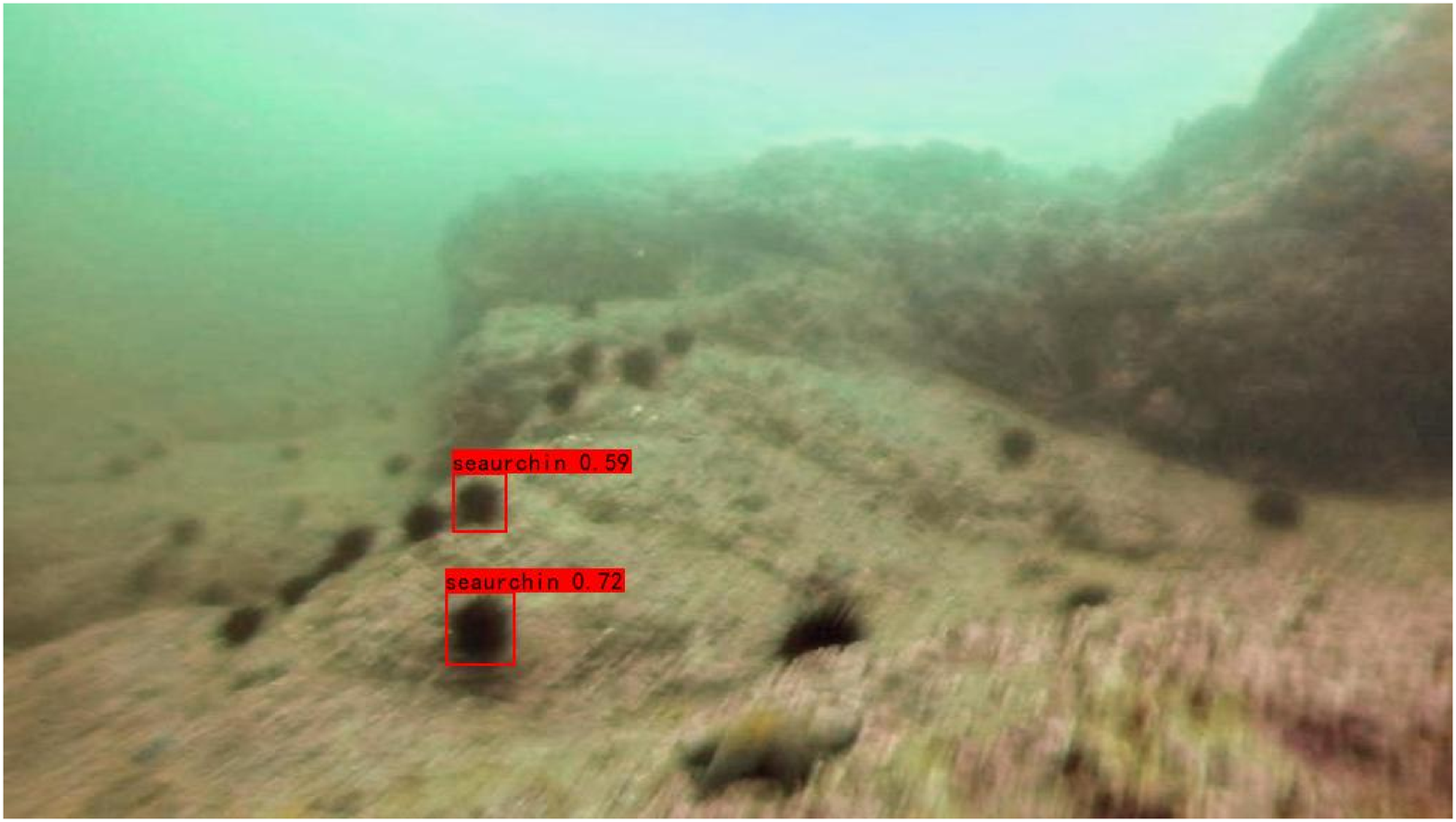}}
		\qquad
		\subfigure[Image utility quality]{
			\label{Fig. 2. sub. 2}
			\includegraphics[width=5.5cm,height=2.6cm]{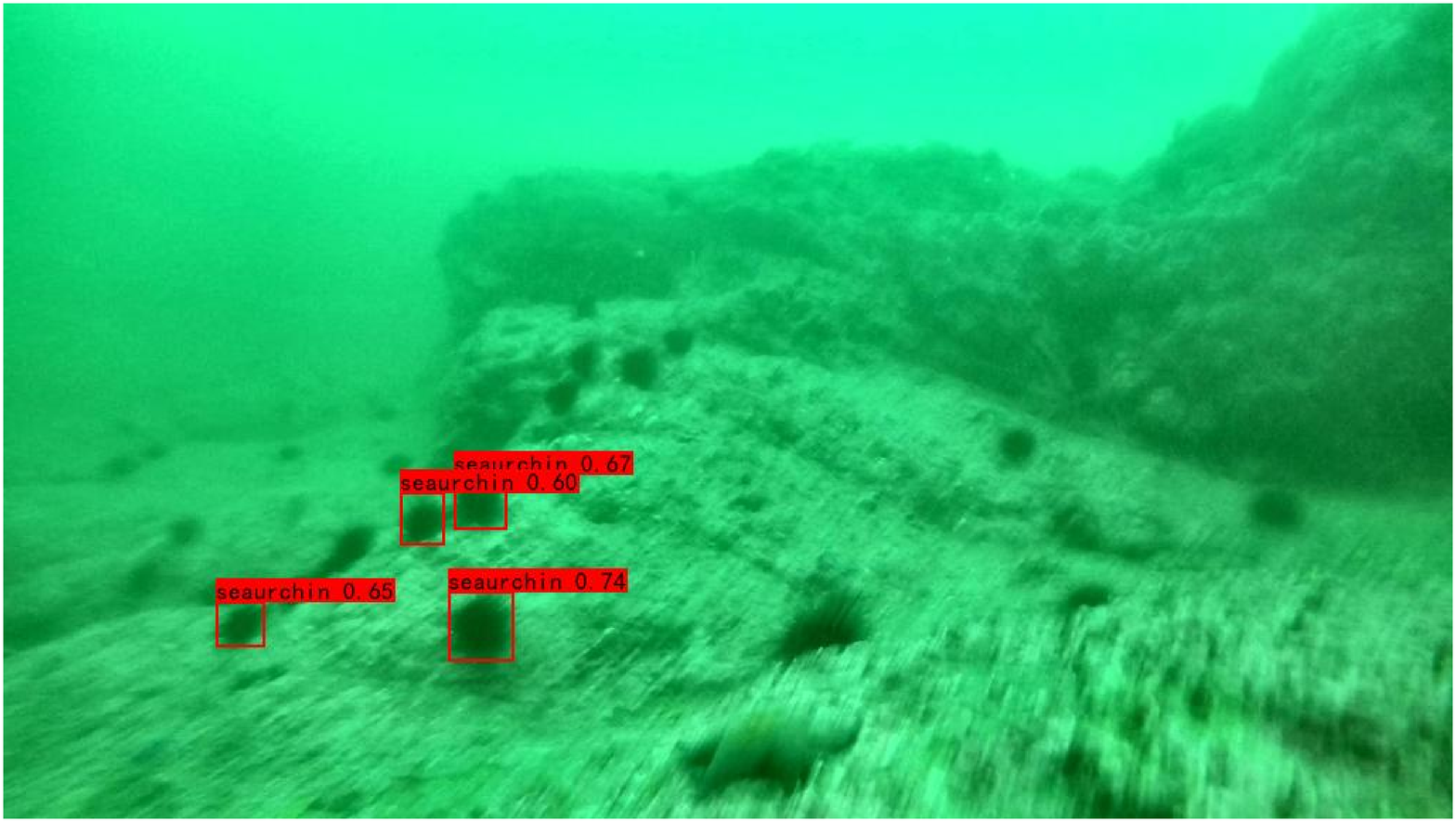}}
		\caption{Comparison on visual to utility.}
		\label{Fig. 2}
	\end{minipage} 
\end{figure}

Underwater object detection, which plays an important role in underwater intelligent applications such as marine ecological monitoring, ocean exploration, and geological mapping, has drawn considerable attention\cite{ref_article1}. However, due to the limits of underwater imaging, the underwater image always appears to be degraded. Fig.\ref{Fig. 1} shows underwater images with typical distortions. Unideal light sources and uneven lightwave attenuation can cause non-uniform illumination and color degradation. Suspended small particles and camera equipment jitter may result in ocean snow noise and motion blur. The negative consequence of these degradation phenomena is reflected in the absence of target contours, the lack of effective details, and the corruption of high-level semantic information. Therefore, underwater images should be enhanced to support subsequent underwater object detection tasks.

In recent years, research on underwater image enhancement is focused on enhancing image visual quality. The research strategies can be broadly classified into two categories: physical model-based and deep learning-based strategies. In the first strategy, researchers usually start with a physical degradation model based on underwater imaging principles, then estimate unknown model parameters through prior assumptions. Researchers investigate various prior assumptions including underwater dark channel prior (UDCP)\cite{ref_proc1} and underwater light attenuation prior (ULAP)\cite{ref_lncs1} and color space dimensionality reduction prior (CSDRP)\cite{ref_proc2}. The second strategy almost relies on the amount of paired training data and the learning ability of deep neural networks. References\cite{ref_proc3} and\cite{ref_article2} synthesized a large number of undistorted underwater images via generative adversarial networks (GAN) to acquire enough data to train enhancement algorithms. Based on massive data, references\cite{ref_article3} and\cite{ref_article4} build a series of end-to-end enhancement networks to learn features from various underwater image domains. These enhancement methods can help to improve image quality towards human commonsense and aesthetics, however, may not always contribute to detection accuracy\cite{ref_proc4r}.

As shown in Fig.~\ref{Fig. 2}, image utility quality is distinct from image visual quality. The richness of effective information, which is helpful to recall and precision of detection task, determines image utility quality. Reference\cite{ref_proc5} states that reducing noise and removing blur are effective approaches to increase classification accuracy. Since low-light images hurt detection performance, reference\cite{ref_article5} develops an unsupervised bio-inspired two-path network (BITPNet) for enhancing nighttime traffic images. As weak contrast and insufficient color may weaken high-level semantic information, references\cite{ref_article6} and\cite{ref_article7} guide the training of the enhancement network through the feedback information provided by a detection model.

Nevertheless, current studies on image utility quality enhancement cannot be generalized to underwater scenes. Different from the atmospheric environment, the imaging conditions of underwater images are severe. These enhancement methods perform poorly in degraded underwater images. Meanwhile, existing works, that are optimized for a fixed classifier or detector only, cannot fully improve the performance of different detectors. The limited capacity of underwater delivery equipment and high real-time requirements for underwater object detection tasks bring great challenges to underwater image utility quality enhancement.

To address these concerns, a task-oriented dynamic enhancement framework based on low-level corrections is proposed to back up detection tasks. Our unique contribution can be summarized as follows:

\begin{itemize}
	\item[$\bullet$] We propose a framework for underwater image utility quality enhancement based on low-level corrections. The utility quality strategy and lightweight corrections can achieve real-time detection performance improvement in diverse underwater applications. 
	\item[$\bullet$] We investigate different contributions of image features to the image utility quality and construct a contribution dictionary based on various detection network architectures. The dynamic method can be widely applied before object detection tasks or other high-level vision tasks. 
\end{itemize}

\section{Method}
\subsection{Low-Level Corrections}
Low-level information distortion will cause changes in high-level semantic information. The primary reason for detector inaccuracy is the degradation of underwater images, including distortions in brightness, color, clarity, and contrast. In the case of low brightness, insufficient luminance may lead to the loss of edge and contour information, making object localization difficult; in the case of excessive brightness or overexposure, detail information is lacking, making accurate object classification difficult. Similarly, color distortion can make it difficult to match the real color features, thus dramatically reducing classification accuracy. Lower clarity and contrast can weaken the high-level semantic information of an image, thus affecting detection performance.

Therefore, this paper focuses on the following low-level corrections, and the detailed operations are shown in Table~\ref{tab1}.

\begin{itemize}
	\item Gamma transformation is performed for brightness distortion. A nonlinear mapping of gray levels is used to achieve brightness equalization. The parameter of gamma transformation is set to 0.5 for stretching gray values in dark areas for low-light underwater images. 
	\item White balance is employed for color distortion. White balance processes the RGB channels of an image to eliminate the effects of underwater environments and restore color information.
	\item Median filtering is performed for clarity distortion. The filter sets each pixel point’s gray value to the median of all pixel points in its neighboring window, enabling image pixel values closer to their true values. Noise impact is relieved by clearing up isolated points.
	\item Contrast-limited adaptive histogram equalization (CLAHE) is used for contrast distortion. CLAHE achieves global contrast enhancement by using the mapping curve determined by the histogram of grayscale distribution.
\end{itemize}

\begin{table}[t]
	\caption{Distortions/Corrections/Features}\label{tab1}
	\begin{center}
		\renewcommand\arraystretch{1.6}
		\begin{tabular}{l|l|l}
			\hline
			Distortion & Correction & Feature \\
			\Xhline{1pt}
			Brightness & Gamma transformation & Brightness \\
			Color & White balance & Saturation \\
			Clarity &  Median filter & Entropy\\
			Contrast & Contrast-limited adaptive histogram equalization & Gradient \\
			\hline
		\end{tabular}
	\end{center}
\end{table}

\begin{figure}[t] 
	\centering  
	\subfigure[Brightness]{
		\label{Fig. 3. sub. 1}
		\includegraphics[width=2.5cm,height=4.5cm]{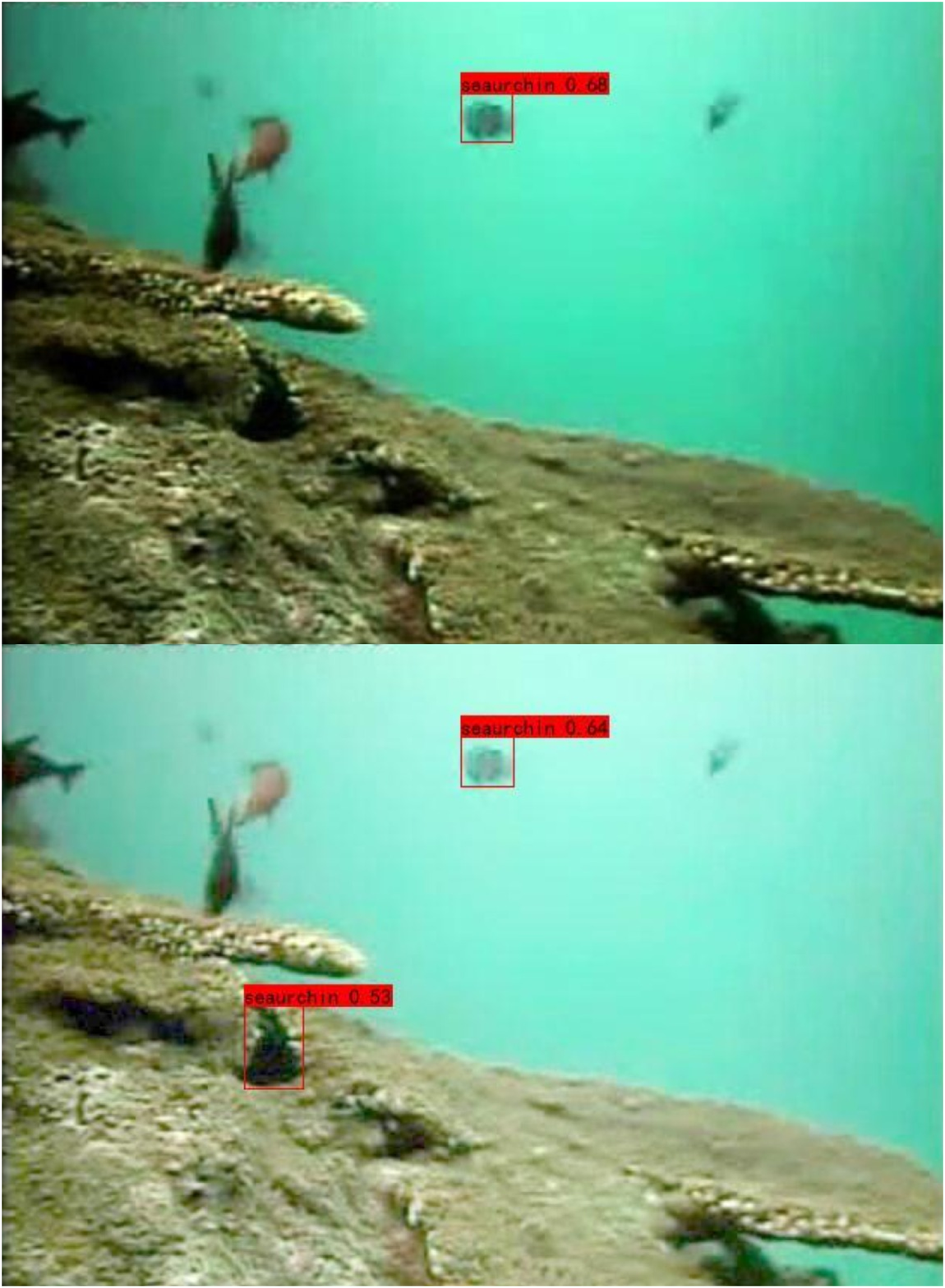}}
	\hspace{-0.5cm}
	\subfigure[Color]{
		\label{Fig. 3 .sub. 2}
		\includegraphics[width=2.5cm,height=4.5cm]{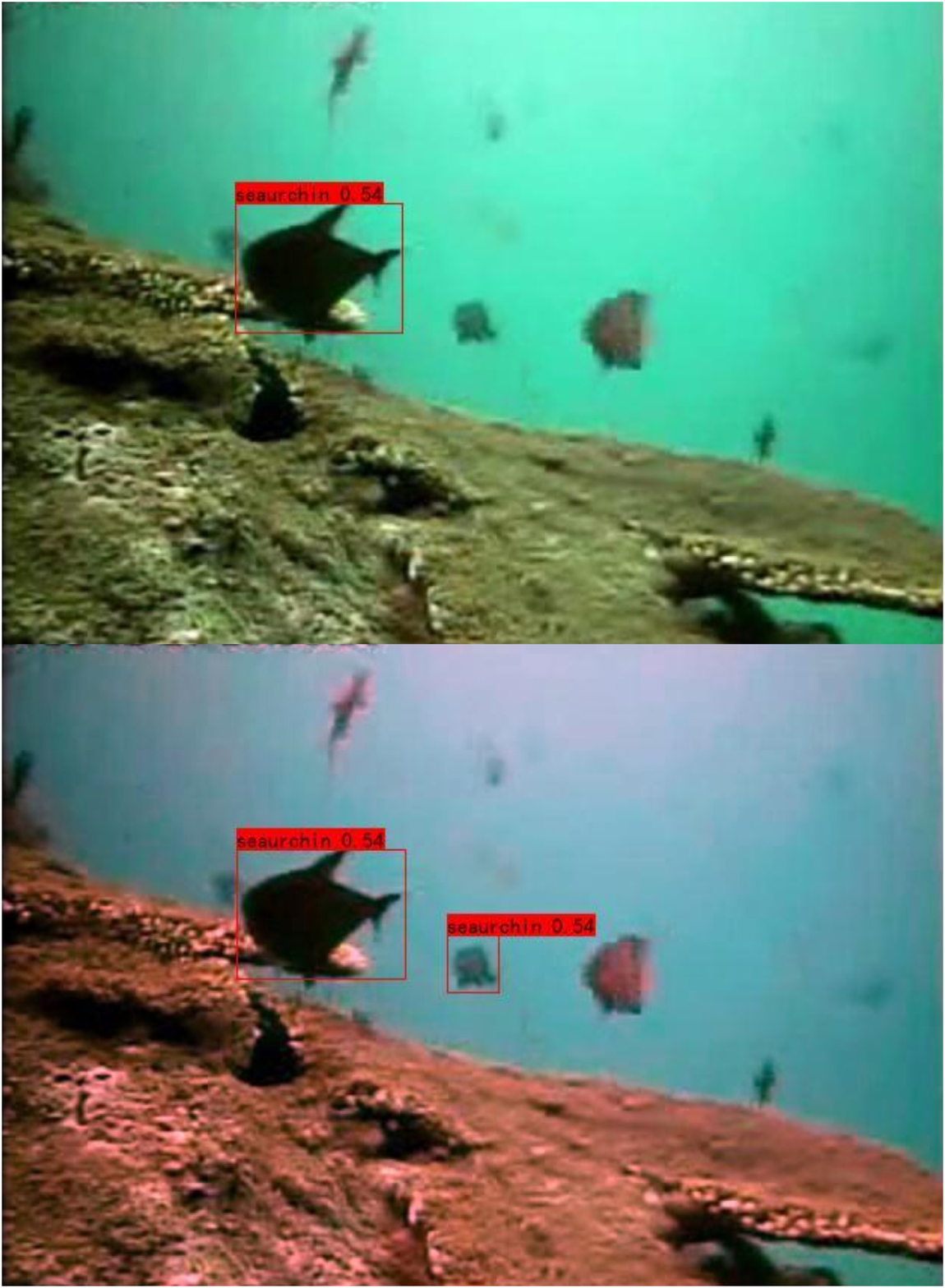}}
	\hspace{-0.5cm}
	\subfigure[Clarity]{
		\label{Fig. 3 .sub. 3}
		\includegraphics[width=2.5cm,height=4.5cm]{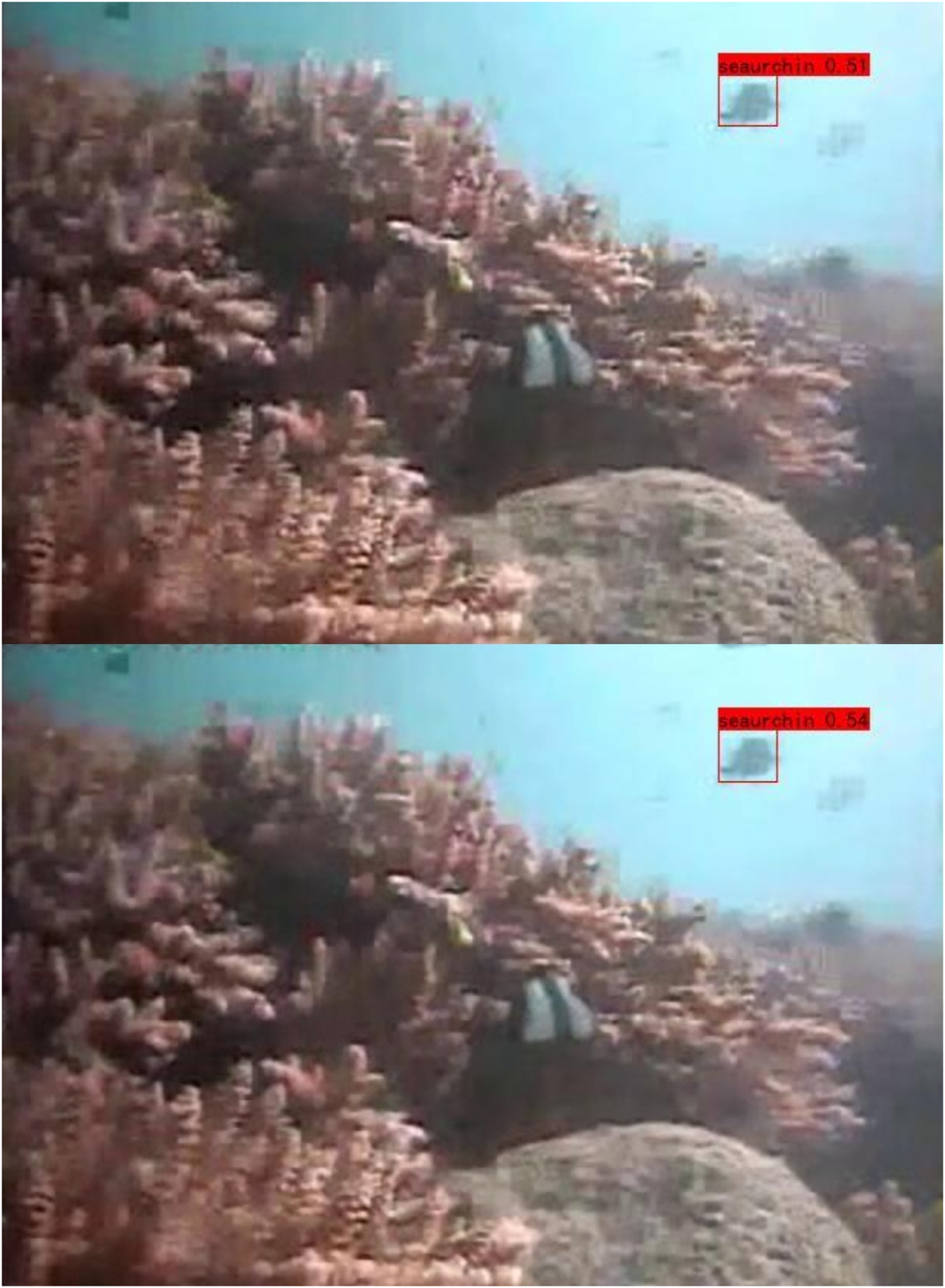}}
	\hspace{-0.5cm}
	\subfigure[Contrast]{
		\label{Fig. 3. sub. 4}
		\includegraphics[width=2.5cm,height=4.5cm]{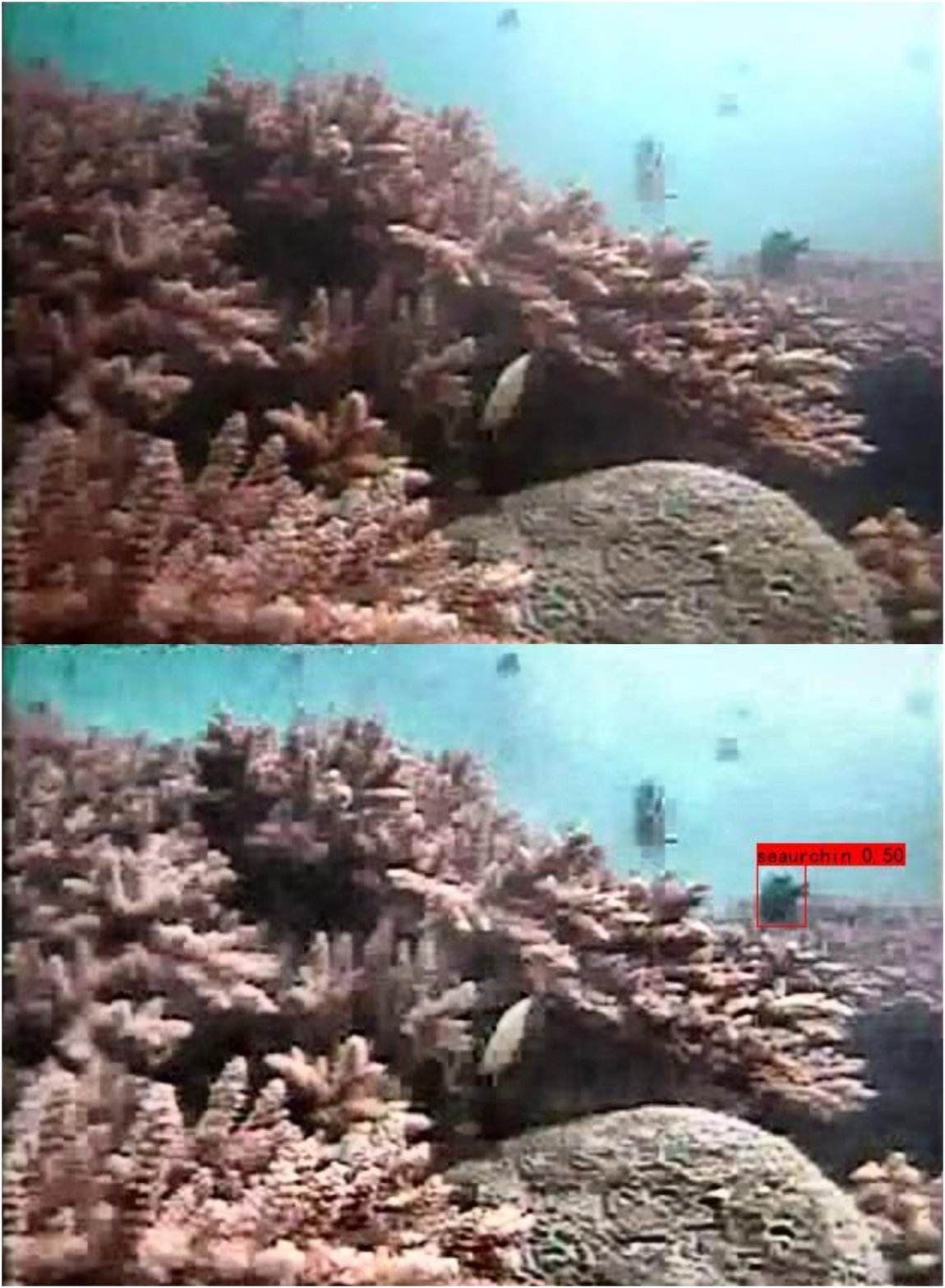}}
	\hspace{-0.5cm}
	\subfigure[Contrast]{
		\label{Fig. 3. sub .5}
		\includegraphics[width=2.5cm,height=4.5cm]{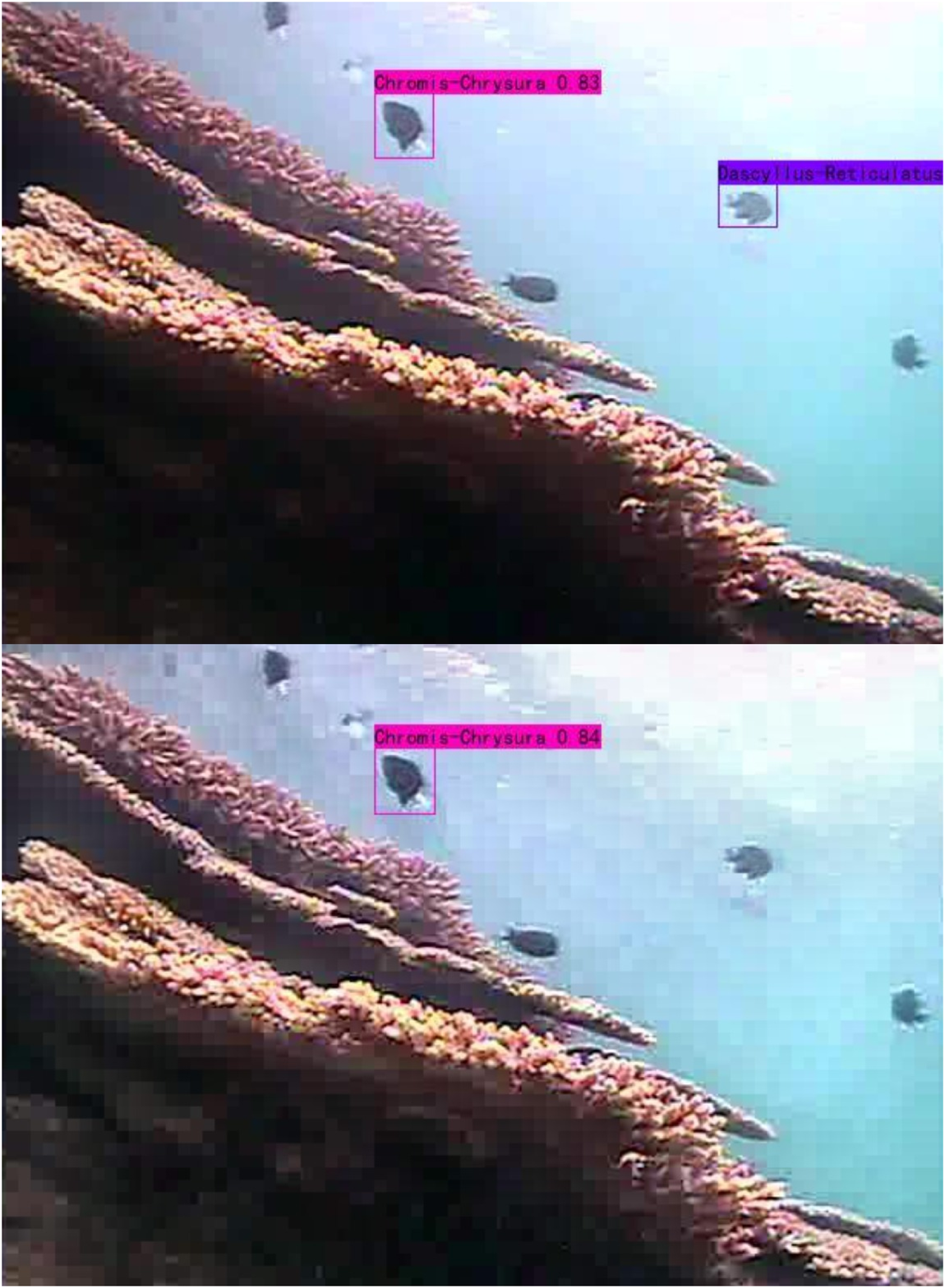}}
	\caption{Results of low-level corrections. Raw images are in the top row, enhanced images are in the bottom row.}
	\label{Fig. 3}
\end{figure}

\noindent{As shown in Fig.~\ref{Fig. 3}, low-level corrections can help to reduce the number of missed objects and increase confidence in correct objects. Compared to the other correction operations, clarity correction only improves the confidence of correct objects which helps a little with detection. Moreover, excessive corrections can instead lead image features to become more unfavorable to be detected. For instance, over-contrast correction introduces additional noise.}

\subsection{Image Utility Quality}\label{s2.2}
In this paper, we use image utility quality to quantify the degree of underwater image distortion.

Image utility quality represents the richness of valid information that is concerned with machine vision tasks. In object detection tasks, the image utility quality of the entire dataset is reflected as the mean average precision (mAP), which is formulated as:

\begin{equation}
mAP =  \dfrac{1}{k}\sum_{i=1}^k\sum_{i=1}^{n-1}(R_{i+1}-R_{i})P_{inter}(R_{i}+1)\label{eq1}
\end{equation}

\noindent{where,}

\begin{equation}
P = \dfrac{TP}{TP+FP}\label{eq2}
\end{equation}

\begin{equation}
R = \dfrac{TP}{TP+FN}\label{eq3}
\end{equation}

\noindent{${TP}$ is the number of true positives, ${FP}$ is the number of false positives, and ${FN}$ is the number of false negatives. ${AP}$ represents the area under the Precision-Recall curves. However, ${mAP}$ is insufficient for describing the utility quality of a single image. Due to the influence of isolated variables, image utility quality cannot be exactly quantified. The Precision-Recall curve fluctuates greatly, and the results of ${mAP}$ are inconsistent in the whole image dataset and a single image. For a more correct definition of image utility quality, detailed elements in the ${mAP}$ equation were introduced to calculate the image utility quality score.}

\begin{equation}
Q = mAP - \dfrac{FN}{GT} + C_{TP} - C_{FP}\label{eq4}
\end{equation}

\noindent{where, ${GT}$ is the number of ground truths. ${C_{TP}}$ denotes the smallest confidence value in the ${TP}$ list and ${C_{FP}}$ denotes the largest confidence value in the ${FP}$ list. Reducing ${FP}$, increasing ${C_{TP}}$, or decreasing ${C_{FP}}$ are effective solutions that can improve image utility quality. With this, each correction can split the image dataset into two categories. One type is required to implement the correction, whereas the other is not required.}

Therefore, low-level corrections that reduced underwater image distortions can improve image features, enhance image utility quality, and boost the performance of detection tasks.

\subsection{Contribution Dictionary}

High real-time performance is required for object detection tasks. Tasks background can be classified into two types: one is the direct application of intelligence to the acquired images in complicated and changing underwater scenes. Underwater equipment, however, has limited carrying capacity for complex computing operations. The other is to apply the underwater images after transmitting them to land via an underwater channel. However, underwater channels suffer from restricted bandwidth, multipath, quick fading, and other defects. The transmission delay makes it impossible to respond to underwater changes promptly. As a result, as a pre-processing step, the image utility quality enhancement must focus on another major indicator of detection performance: time complexity. It is essential to balance the improvement contribution and time consumption introduced by correction operations.

\begin{figure}[t]
		\begin{minipage}{0.5\textwidth}
			\centering
		    \makeatletter\def\@captype{table}\makeatother
			\caption{Popular detector}
			\label{tab2}
			\begin{center}
			\renewcommand\arraystretch{1.6}
			\begin{tabular}{l|l}
				\hline
				Method & Backbone \\
				\Xhline{1pt}
				SSD\cite{ref_lncs2} & VGG-16 \\  
				Efficientdet\cite{ref_proc6}& EfficientNet-B0 \\ 
				Centernet\cite{ref_proc7}& ResNet-50 \\  
				YOLOX\cite{ref_article8} & CSPDarknet  \\  
				\hline
			\end{tabular}
		\end{center}
		\end{minipage}
		\hspace{-0.3cm} 
		\begin{minipage}{0.5\textwidth}
			\centering
			\subfigure[Centernet]{
				\label{Fig. 4. sub. 1}
				\includegraphics[width=2.5cm,height=2.5cm]{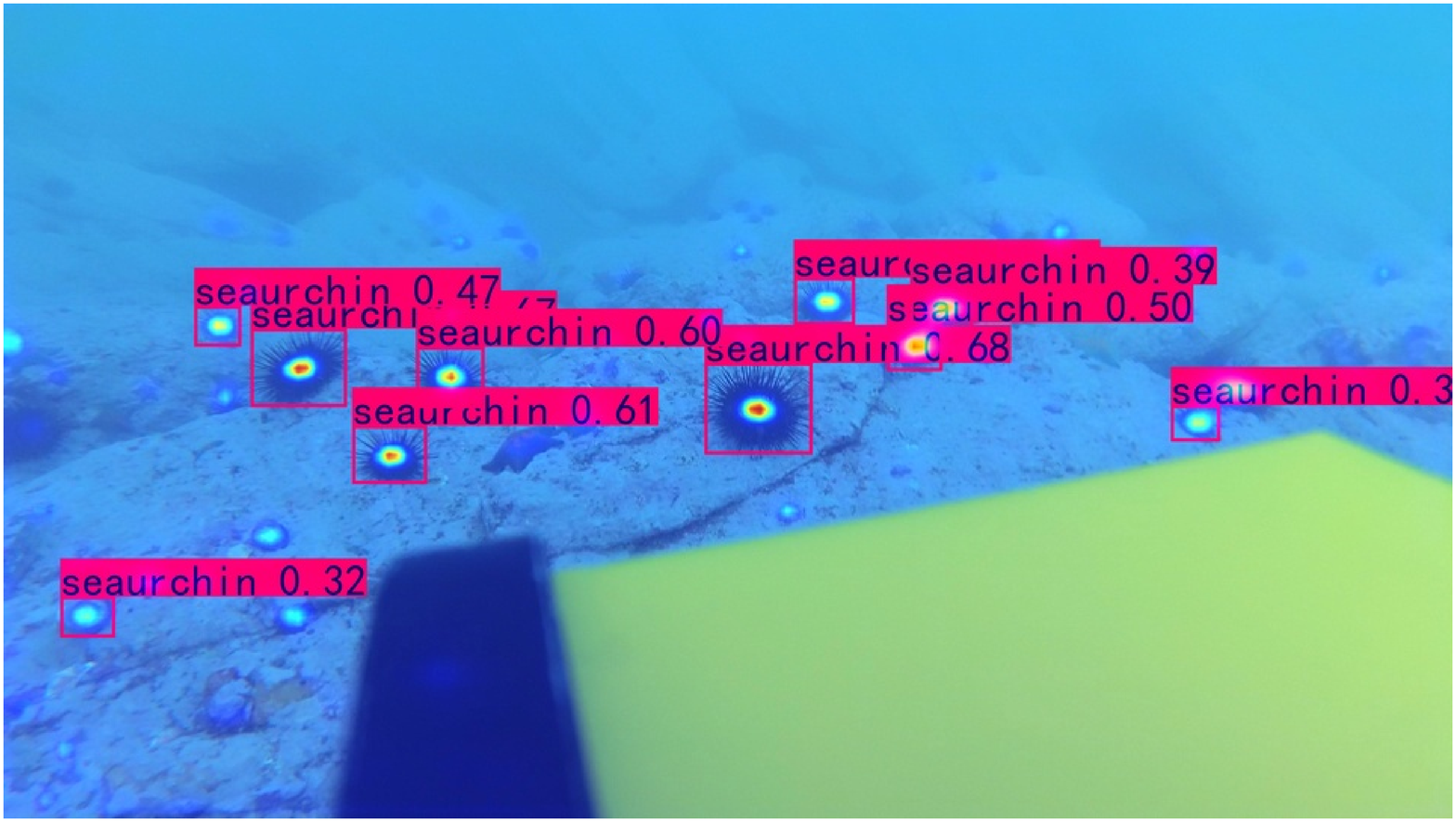}}
			\hspace{-0.3cm} 
			\subfigure[YOLOX]{
				\label{Fig. 4. sub. 2}
				\includegraphics[width=2.5cm,height=2.5cm]{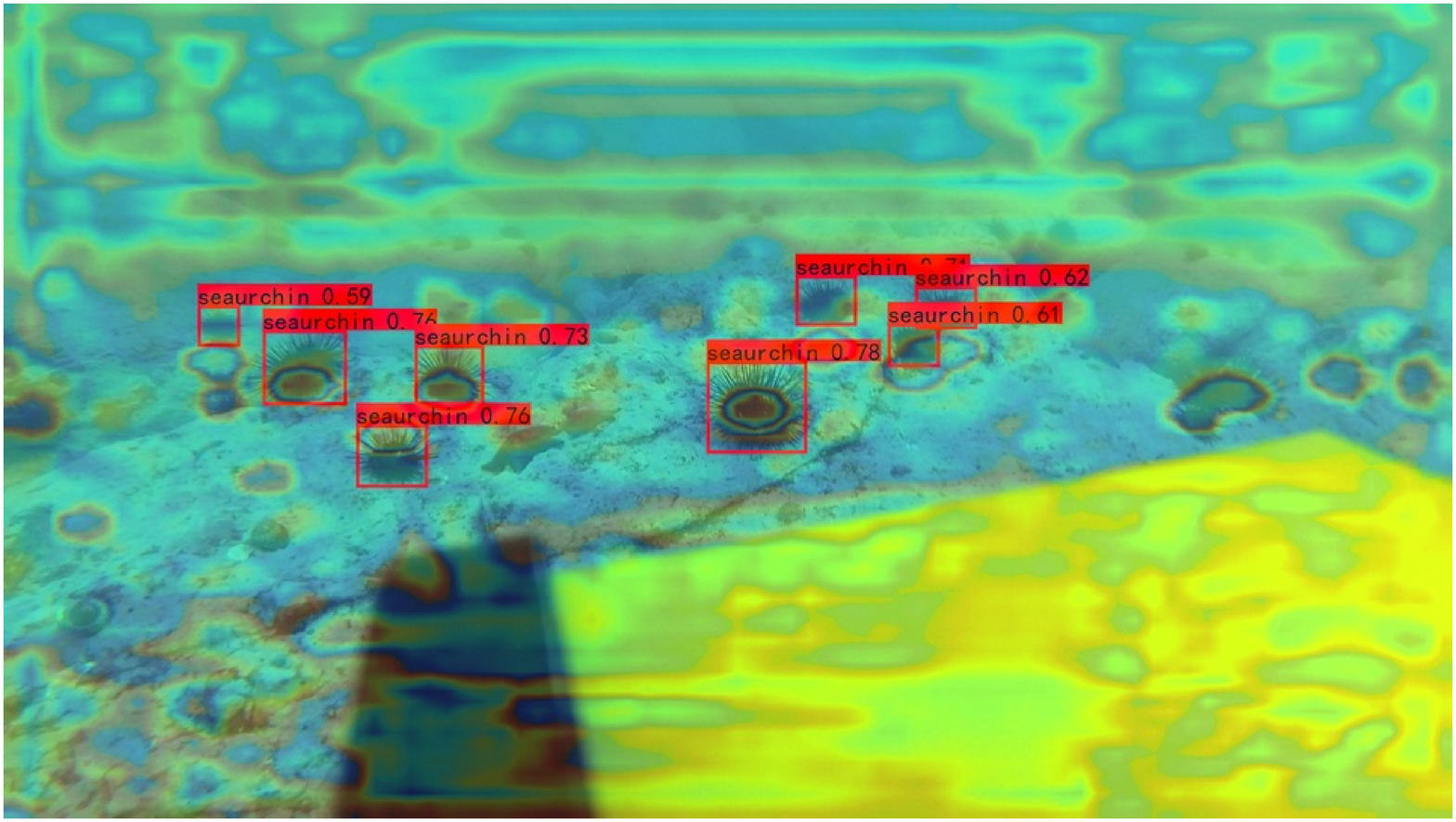}}	
			\caption{Prediction feature maps.}
			\label{Fig. 4}
		\end{minipage}
\end{figure}

Table~\ref{tab2} shows the popular object detection networks for underwater tasks. For feature extraction, the network backbone varies. For feature matching, the anchor strategy differs. In general, for the anchor-free strategy, the predicted feature maps with key points focus more on global features. For shallower backbones, with the small respective field, will extract more fine-grained features from an image. As the number of downsampling or convolution increases, the respective field gradually expands, so more high-level semantic features can be extracted to form the final prediction feature map. The prediction feature maps of different detectors are shown in Fig.~\ref{Fig. 4}.

Even for the same image, utility quality can vary dramatically due to the differences in image features used. Equation~\ref{eq4} also supports this conclusion. Since detectors pay different attention to image features, this paper tends to construct a contribution dictionary for indicting the correlation between image features and image utility quality. The contribution dictionary can be determined as follows:  

\begin{equation}
I = \omega*\xi(C,D)\label{eq5}
\end{equation}

\noindent{where, ${\omega}$ refers to the contribution label corresponding to correction operation. When the image feature value matches the applicable range, the contribution label is set to 1. Otherwise, set the contribution label to 0. ${\xi}$ refers to the contribution weight of this correction operation. The contribution weight is the normalized correlation coefficient. The stronger the correlation between image feature and image utility quality, the higher the weight. Hence, the contribution dictionary can be determined by the low-level corrections: ${C}$ and the detector: ${D}$.}

In general, ${\xi}$ of each correction is ranked as contrast correction, color correction, clarity correction, and brightness correction. According to this order, the enhancement framework first cascades correction operations.

\begin{equation}
B = \dfrac{I}{T}\label{eq6}
\end{equation}

\noindent{${I}$ is the contribution gains calculated from Equation~\ref{eq5}. ${T}$ is the time complexity obtained from prior knowledge. Since gradient always be a key feature in image utility quality, the initial benchmark of ${B}$ value is caculated by contrast correction. When the new correction operation gets a higher ${B}$ value, the operation will be performed and its parameters will be updated to be a new benchmark later. Finally, an optimal combination of corrections can be generated.}

\subsection{The Overview of The Proposed Framework}

\begin{figure}[t]
	\centering
	\includegraphics[width=\textwidth]{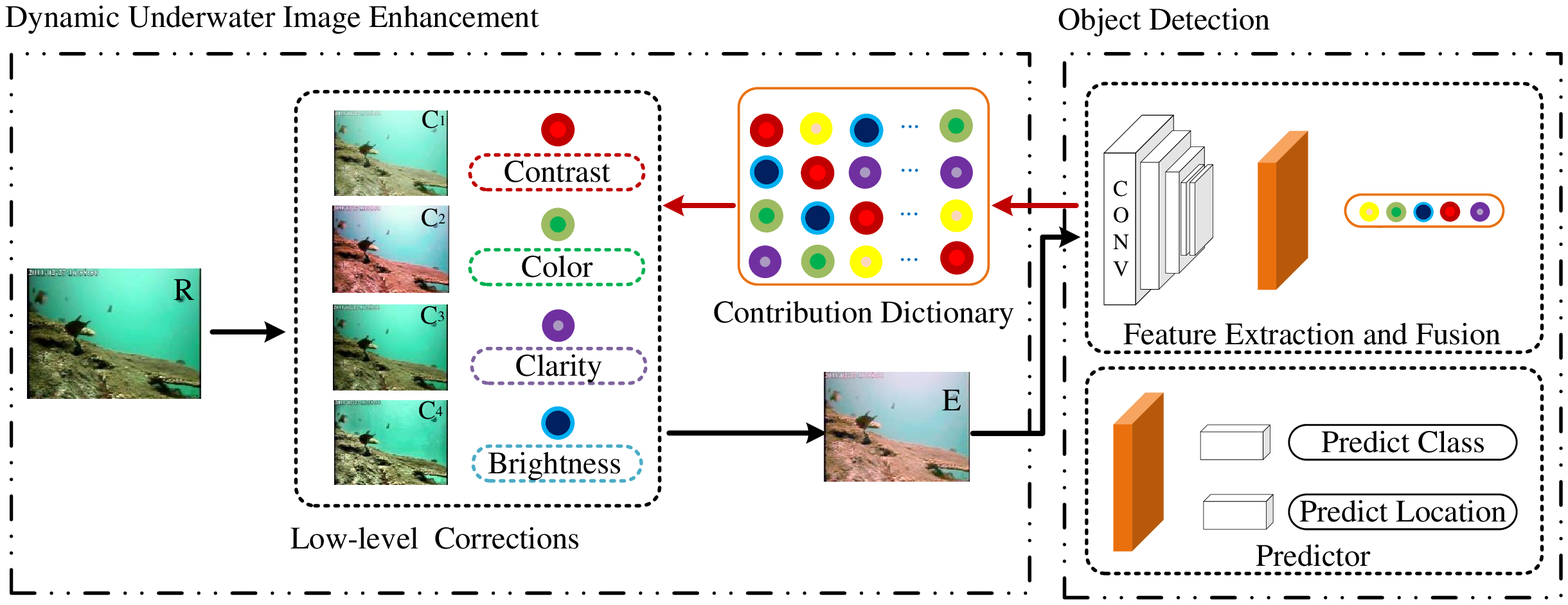}
	\caption{Image utility quality enhancement framework for underwater object detection.} \label{Fig. 5}
\end{figure}

The underwater image utility quality enhancement framework proposed in this paper is shown in Fig.~\ref{Fig. 5}. The implementation of the dynamic low-level corrections is guided by a contribution dictionary based on detection prior. The original underwater images are first pre-processed by the image enhancement module to enhance image utility quality and generate enhanced images before being sent into the object detection module. In the image enhancement module, according to the analysis of the underwater degradation, low-level correction operations consist of contrast correction, color correction, clarity correction, and brightness correction. In the object detection module, the detector does not refer to a fixed detection model, but to detection models used in specific tasks.

Our method is task-oriented, aiming to improve the image utility quality. With the work above, the framework provides dynamic and optimal enhancement paths for different detection tasks, that can be generalized in underwater scenes.

\section{Experiments}

\subsection{Implementation Details}

The UDD dataset\cite{ref_article9} is an underwater image dataset for robot target capture tasks, that was created by the Dalian University of Technology underwater robotics team. It contains three categories: sea urchin, sea cucumber, and scallops, with a training set of 1827 images and a testing set of 400 images. Original images suffered from severe motion blurring and color degradation. In the following discussion, only sea urchins will be evaluated.

The Fish4knowledge dataset is derived from an official SeaCLEF competition. It contains eight species of fish: Chaetodon lunulatus, Pempheris vanicolensis, Amphiprion clarkii, Chaetodon trifascialis, Chromis chrysura, Dascyllus reticulatus, Plectrogly phidodon dickii and Dascyllus aruanus. The dataset provides images with 640$\times$480 or 320$\times$240 pixels. The training set covers 6436 images and validation set covers 1462 images and test set covers 2742 images. 

Table~\ref{tab3} shows equations for image features corresponding to corrections. As described in Section~\ref{s2.2}, we can obtain ${\omega}$ by calculating feature values.

\begin{table}[t]
	\caption{Feature/Calculation Formula/Value}\label{tab3}
	\begin{center}
		\renewcommand\arraystretch{1.8}
		\begin{tabular}{l|l|l}
			\hline
			Feature & Calculation Formula & Value \\
			\Xhline{1pt}
			Gradient & 
			$Gradient=\dfrac{1}{x*y}\sum_{x}\sum_{y}(G_x*I(x,y)+G_y*I(x,y))^{2}$ & 0 - 0.9 \\
			Saturation & $Saturation=({max(R,G,B)-min(R,G,B)})/{max(R,G,B)}$ & 0.3 - 0.5 \\
			Entropy &  
			$Entropy=-\sum_{0}^{255}{P_{i}log_{2}P_{i}}$ & 0 - 0.9\\
			Brightness & 
			$Brightness=0.299*R+0.587*G+0.114*B$ & 0.4 - 0.6 \\
			\hline
		\end{tabular}
	\end{center}
\end{table}

\noindent{Specifically, the gradient value can reflect image sharpness and texture variation. The Tenengrad gradient function, where ${G_x}$ and ${G_y}$ are the kernels of the Sobel operators, employs the Sobel operators to extract gradient values in the horizontal and vertical directions, respectively. Image entropy represents the average amount of information in an image. In the calculation formula, ${P_i}$ shows the proportion of pixels in the image with the gray value ${i}$. The entropy and gradient values can be increased by clarity correction and contrast correction, to enhance image utility quality by effectively restoring edges and contour information. Brightness and saturation can be characterized by a linear combination of three elements in the RGB color space. Brightness correction can properly enhance the overall image without introducing additional unfavorable information. Color correction can correct color detail information in the image.} 

Since the YOLOX model and the Centernet model perform well in accuracy and real-time performance among popular detection networks. Meanwhile, there are significant differences between the two models in the feature-extracting and feature-matching process. We carry out experiments based on these two models as illustrations.

According to Equation~\ref{eq4}, the image utility quality score corresponding to the detectors is obtained. And according to the formula in Table~\ref{tab3}, image feature values are acquired. In this part, ${s}$ and ${p}$ here refers to the image utility quality score and image feature values. ${\xi}$ denotes the correlation between utility quality and image features, which is measured according to the following formula:

\begin{equation}
PLCC=\dfrac{\sum_{i=1}^{N}(s_{i}-\overline{s})(p_{i}-\overline{p})}{\sqrt{(\sum_{i=1}^{N}{{(s_{i}-\overline{s})}^2}\sum_{i=1}^{N}{{(p_{i}-\overline{p})}^2})}}\label{eq7}
\end{equation}

\noindent{As a result, the contribution dictionary weights for features changed by corrections and detectors can be recorded in Table~\ref{tab4}.}
			
\begin{figure}[t]
	\centering
		\begin{minipage}{0.5\textwidth}
			\centering
			\makeatletter\def\@captype{table}\makeatother
			\caption{Contribution weight: ${\xi}$${(C, D)}$}
			\label{tab4}
			\begin{center}
			\renewcommand\arraystretch{1.6}
			\begin{tabular}{l|l|l}
				\hline
				Feature &  YOLOX & Centernet \\
				\Xhline{1pt} 
				Gradient& 0.4229 & 0.3707 \\  
				Saturation& 0.3768& 0.2808 \\  
				Brightness & 0.3222 & 0.2810 \\ 
				Entropy& 0.1073 & 0.0933 \\  
				\hline
			\end{tabular}
		\end{center}
		\end{minipage}
		\hspace{-0.3cm}  
		\begin{minipage}{0.5\textwidth}
			\centering
			\renewcommand\arraystretch{1.6}
			\makeatletter\def\@captype{table}\makeatother
			\caption{Time complexity: ${T}$}
			\label{tab5}
			\begin{center}
			\begin{tabular}{l|l}
				\hline
				Correction & Time \\
				\Xhline{1pt}
				Contrast& 0.027 \\  
				Color& 0.033 \\  
				Brightness & 0.024 \\
				Clarity& 0.021 \\   
				\hline
			\end{tabular}
		\end{center}
		\end{minipage}
\end{figure}

To measure the time complexity of each correction operation, this paper tests the corrections with the same image resolution and at the same computing device. Table~\ref{tab5} presents the time complexity of each correction.

After the above work, the ${B}$ value can be easily obtained according to Equation~\ref{eq6}, and then the enhancement method can be finally settled down.

The backbone network CSPDarknet of the YOLOX network is a modified Darknet-53, which divides the input feature map into two channels, one of which is connected to the transition layer by residual convolution and 1x1 convolution, the other of which is connected directly to the transition layer by shortcut. More high-level features are extracted as useful features by the YOLOX model. Therefore, the detector is more concerned with gradient features overall. And the dynamic enhancement method is mainly constructed by contrast correction. The Centernet's backbone is ResNet-50. it adopts an anchor-free strategy, estimating the object's location by key points. Therefore, the Centernet model focuses on gradient features as well as saturation and brightness features. Its enhancement method automatically matches contrast correction and brightness correction and color correction cascades.

\subsection{Performance Analysis}
\begin{figure}[t]
		\centering  
		\subfigure[UDD]{
		\includegraphics[width=5cm,height=4cm]{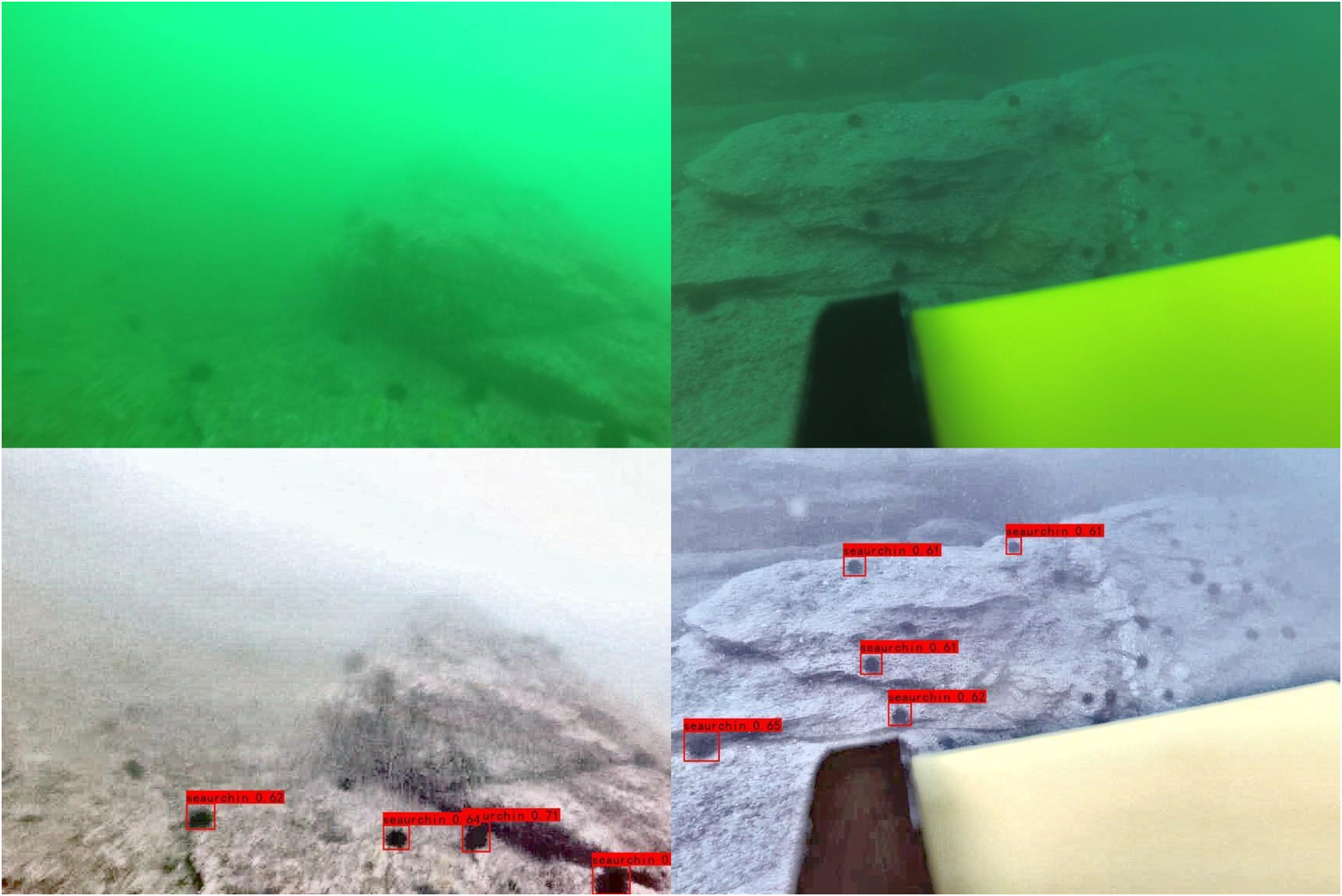}}
		\subfigure[Fish4knowledge]{
		\includegraphics[width=5cm,height=4cm]{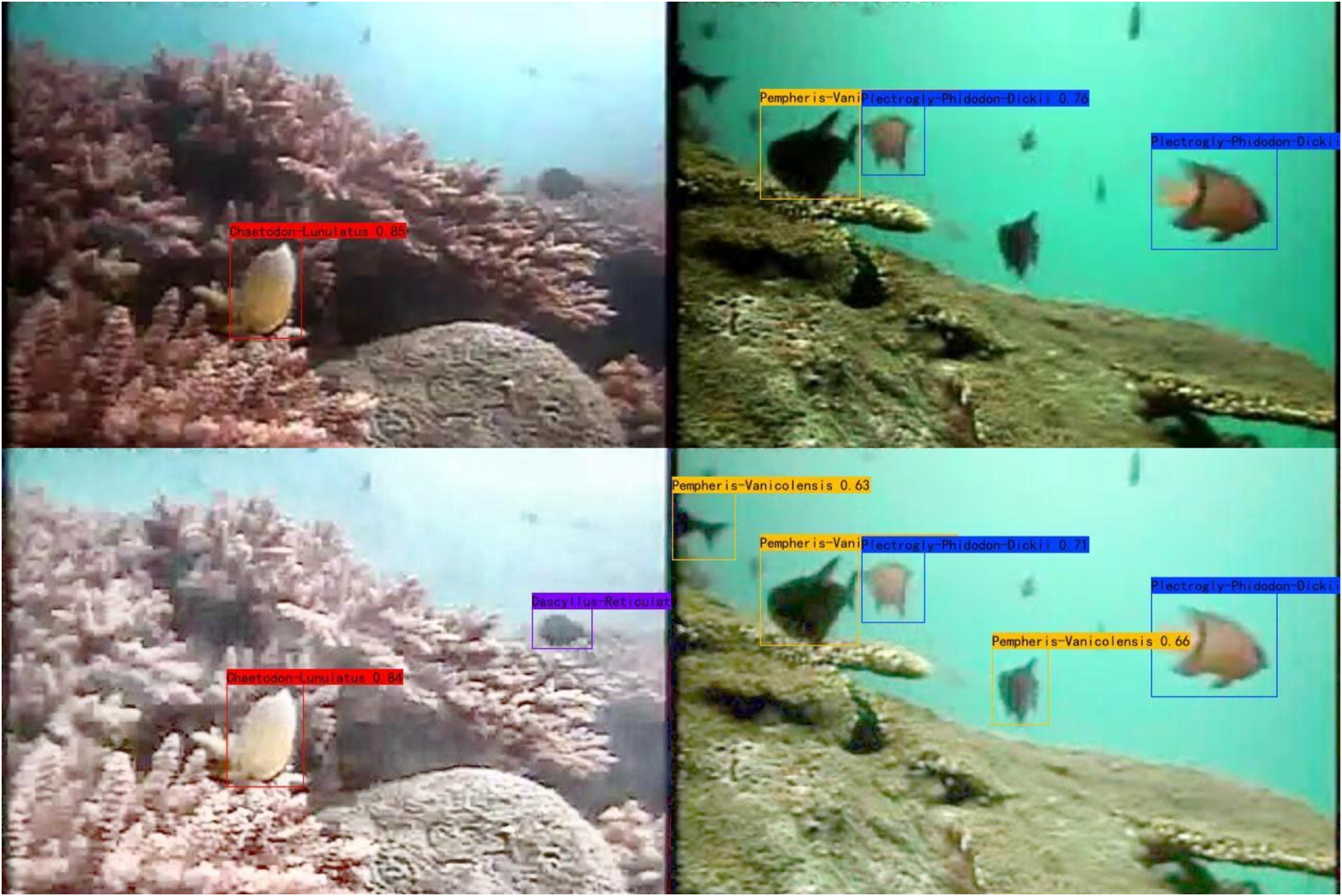}}
		\caption{Underwater object detection results based on our enhancement method. Raw images are in the top row, enhanced images are in the bottom row.}
  \label{Fig. 6}
\end{figure}

We conducted experiments on both UDD and Fish4knowledge for the same detection task. Fig.~\ref{Fig. 6} shows the enhanced results.

\begin{figure}[t]
	\centering  
	\subfigure[UDD]{
		\includegraphics[width=4.5cm,height=4.5cm]{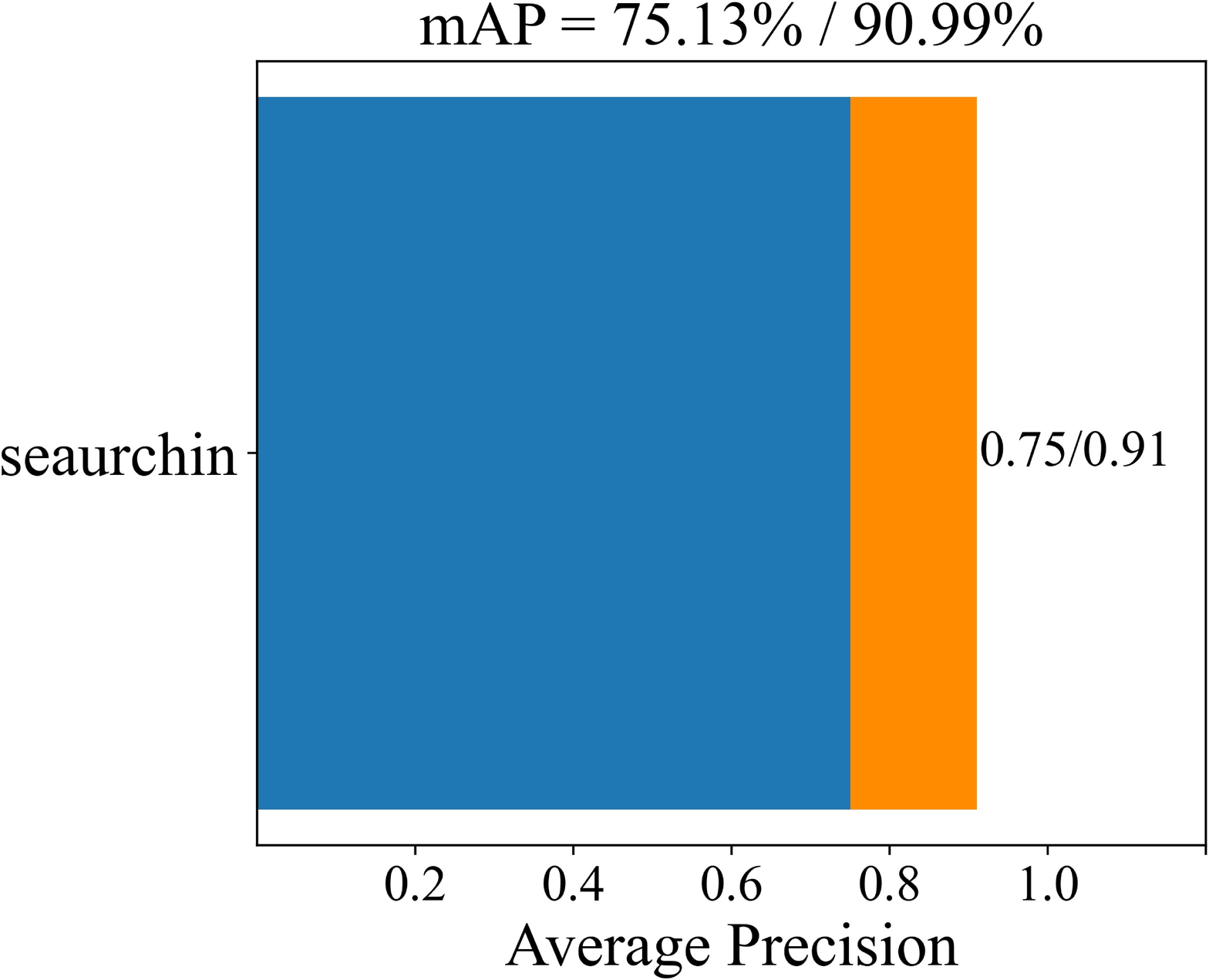}}
	\subfigure[Fish4knowledge]{
		\includegraphics[width=5.5cm,height=4.5cm]{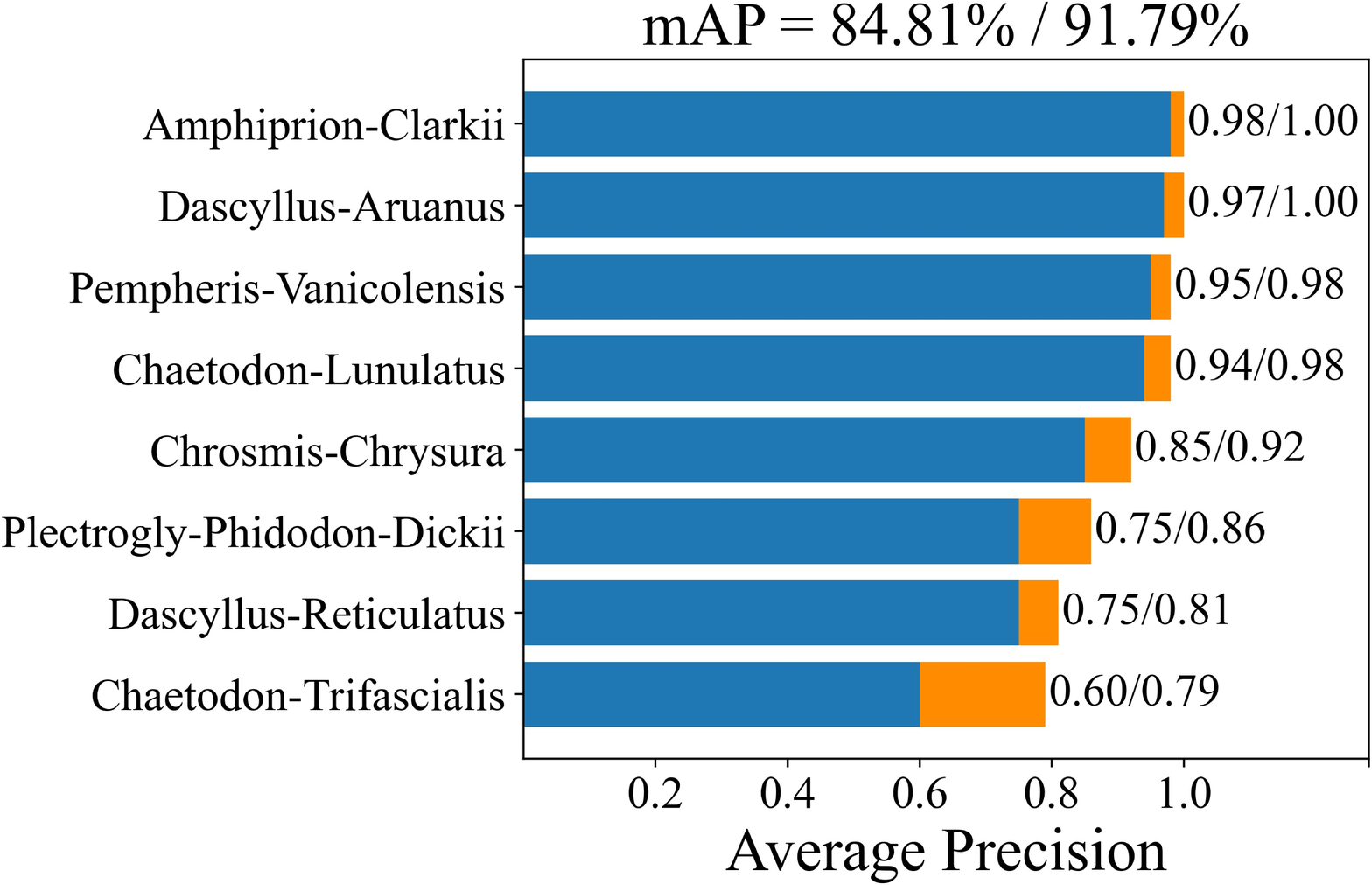}}
	\caption{Underwater object detection performances based on our enhancement method}
	\label{Fig. 7}
\end{figure}

\noindent{After image utility quality enhancement, the subsequent detection can be greatly improved. Enhanced images show a 15.86$\% $/6.98$\% $ improvement in detection accuracy than original images at a speed of 105/126 frames per second (FPS), respectively. Fig.~\ref{Fig. 7} shows the increased result for each category.}

\begin{figure}[t]
 	\centering
	\begin{center}
		\begin{minipage}{0.55\textwidth}
			\centering
			\subfigure[raw]{
				\label{Fig. 8. sub. 1}
				\includegraphics[width=2cm,height=3cm]{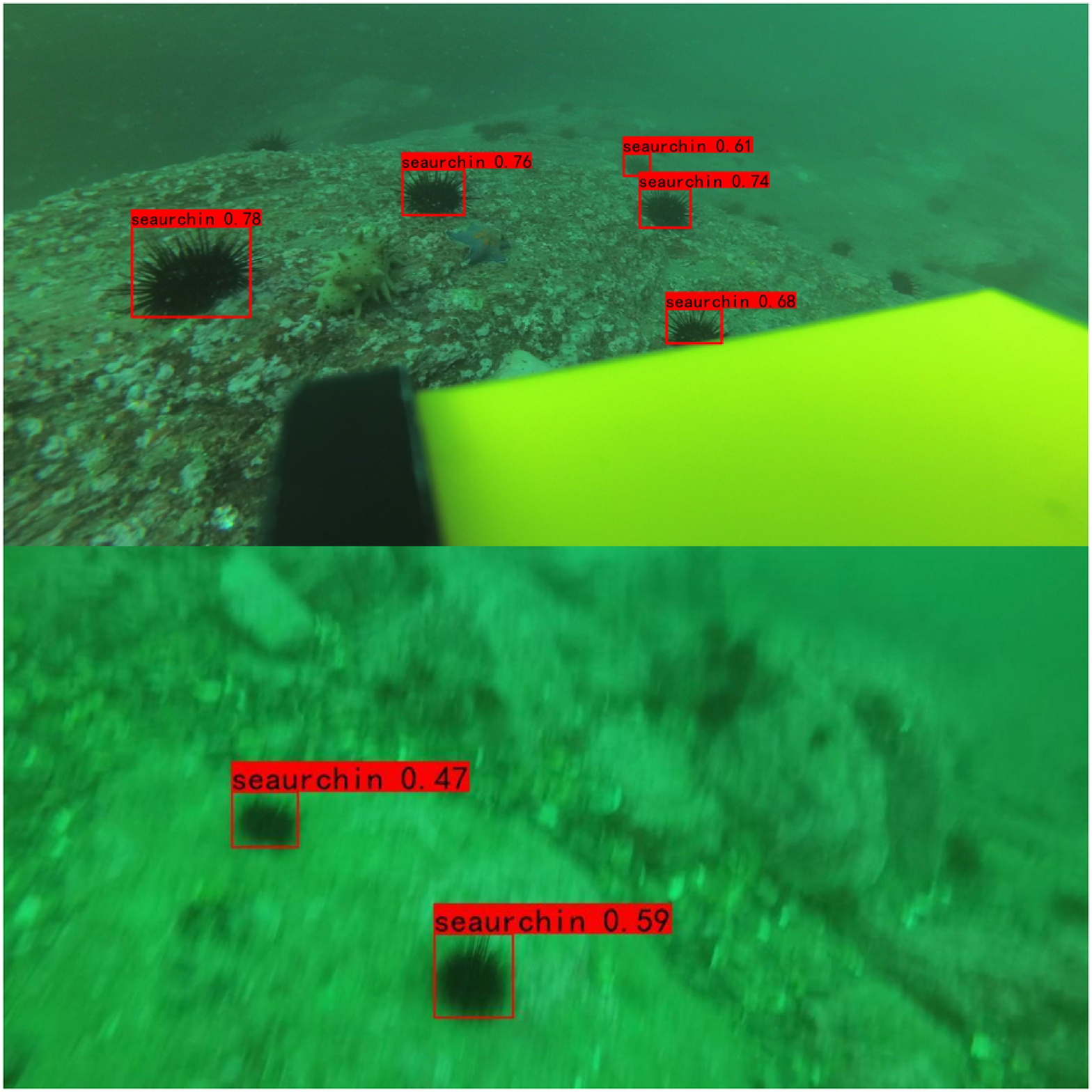}}
			\hspace{-0.3cm}
			\subfigure[Ours$_{1}$]{
				\label{Fig. 8. sub. 2}
				\includegraphics[width=2cm,height=3cm]{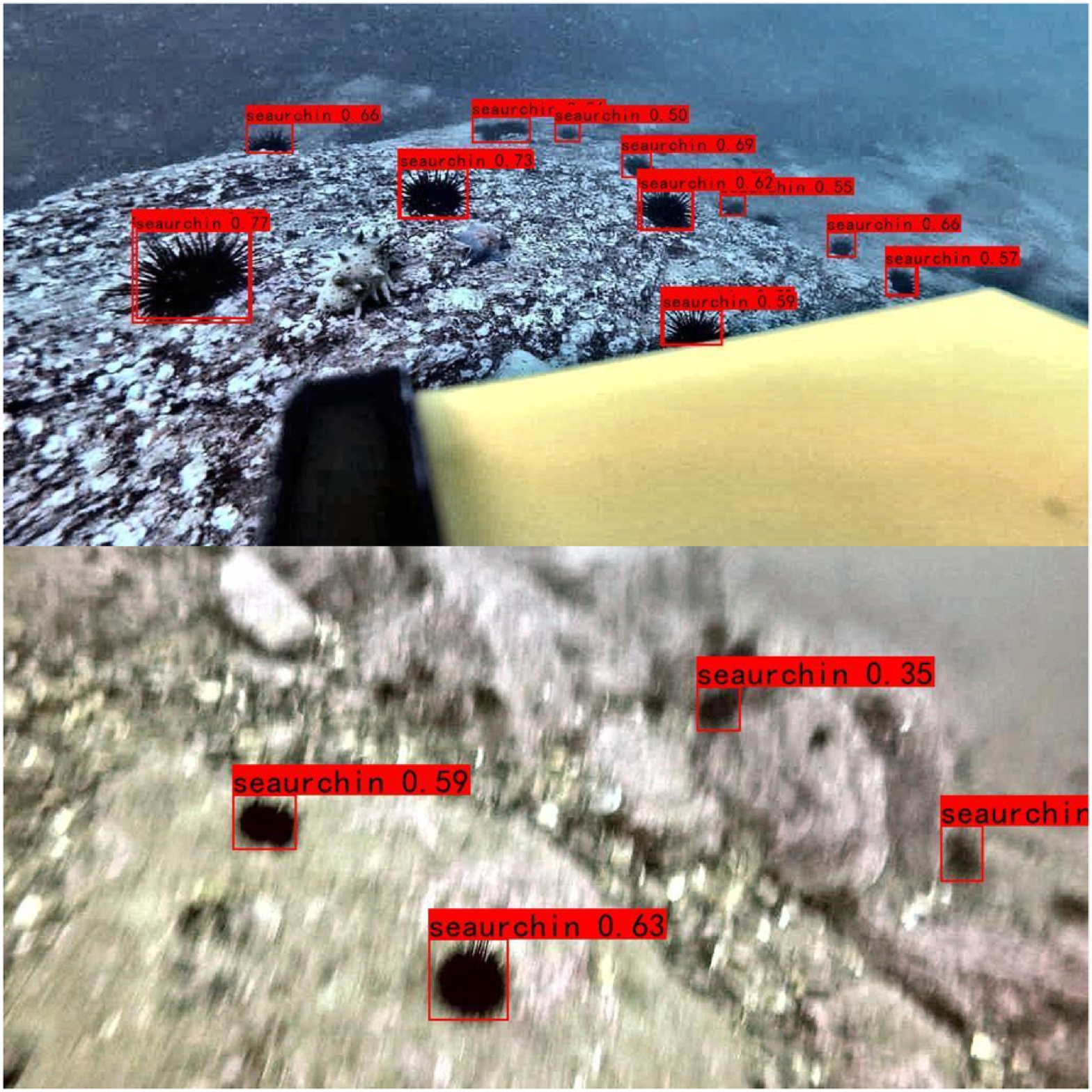}}
			\hspace{-0.3cm}
			\subfigure[Ours$_{2}$]{
				\label{Fig. 8. sub. 3}
				\includegraphics[width=2cm,height=3cm]{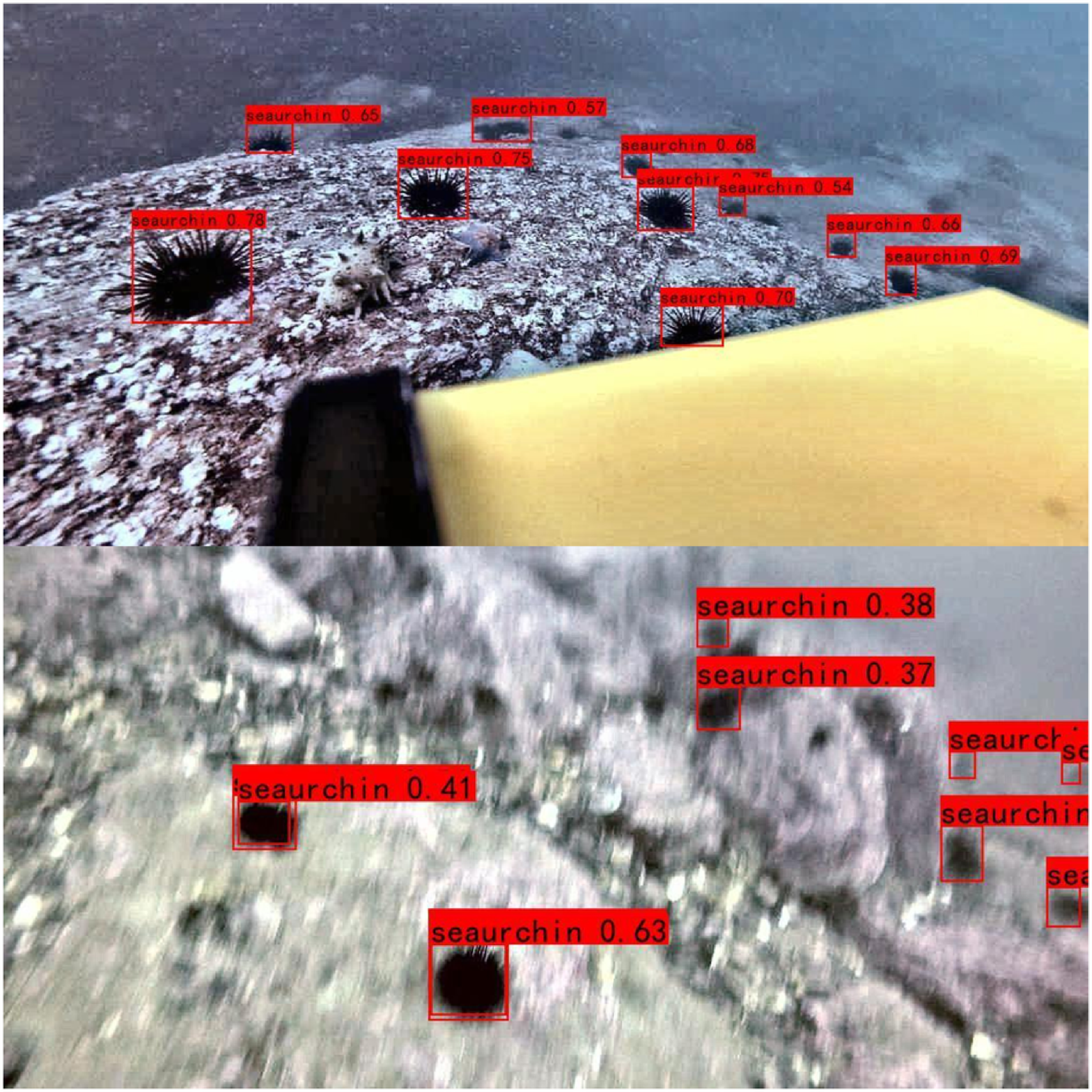}}	
			\caption{Qualitative results comparision.}
			\label{Fig. 8}
		\end{minipage}
		\hspace{-0.3cm}
		\begin{minipage}{0.45\textwidth}
		\centering
		\makeatletter\def\@captype{table}\makeatother
		\caption{Quantitative performances comparision}
		\label{tab6}
		\begin{center}
		\renewcommand\arraystretch{1.6}
		\begin{tabular}{l|l|l|l}
			\hline
			Method & mAP$_{1}$ & mAP$_{2}$ & FPS \\
			\Xhline{1pt}
			raw & 75.13 & 71.83 &  \\  
		    Ours$_{1}$ & \textbf{90.99} &  & 105\\ 
			Ours$_{2}$ & & \textbf{91.22} & 101\\  
			\hline
		\end{tabular}
	\end{center}
	\end{minipage}
	\end{center}
\end{figure}

Furthermore, to verify that the dynamic enhancement method proposed in this paper can provide suitable solutions for various underwater object detection tasks, we train two detection models on UDD. Ours$_{1}$ is designed for the YOLOX model, and mAP$_{1}$ refers to the detection accuracy towards the YOLOX model; Ours$_{2}$ is designed for the Centernet model, and mAP$_{2}$ refers to the detection accuracy towards the Centernet model. The qualitative and quantitative analysis of our methods is shown in Fig.~\ref{Fig. 8} and Table~\ref{tab6}.

Our method provides different correction solutions for different detection tasks and therefore shows the most favorable improvement in detection tasks.
Ablation experiments were conducted to study the contribution of low-level corrections to detection performance. Only one correction operation is carried out for each experiment. Performance comparisons are given in Table~\ref{tab7}.

\begin{table}[t]
	\caption{Ablation performance}
	\begin{center}
		\renewcommand\arraystretch{1.8}
		\begin{tabular}{l|l|l|l|l|l}
			\hline
			Correction & Contrast & Color & Clarity &  Brightness& Ours \\ 
			\Xhline{1pt}
			mAP$_{1}$/FPS & 89.82/126& 87.79/63 & 87.66/188 & 87.74/157  & 90.99/105 \\  
			mAP$_{2}$/FPS & 89.09/100 & 87.73/67 & 87.90/ 175 & 87.92/138 & 91.22/101 \\
			\hline
		\end{tabular}
		\label{tab7}
	\end{center}
\end{table}

The results of the table show that the four aspects discussed in this paper are not redundant. In the four aspects of correction operations, contrast correction seems to be able to make the greatest contribution to the improvement of image utility quality while maintaining a low time complexity.

\subsection{Comparison With State-of-The-Art Methods }
\begin{figure}[t] 
	\centering  
	\subfigure[\fontsize{7pt}{\baselineskip}\selectfont{Raw}]{
		\label{Fig. 9. sub. 1}
		\includegraphics[width=2cm,height=3cm]{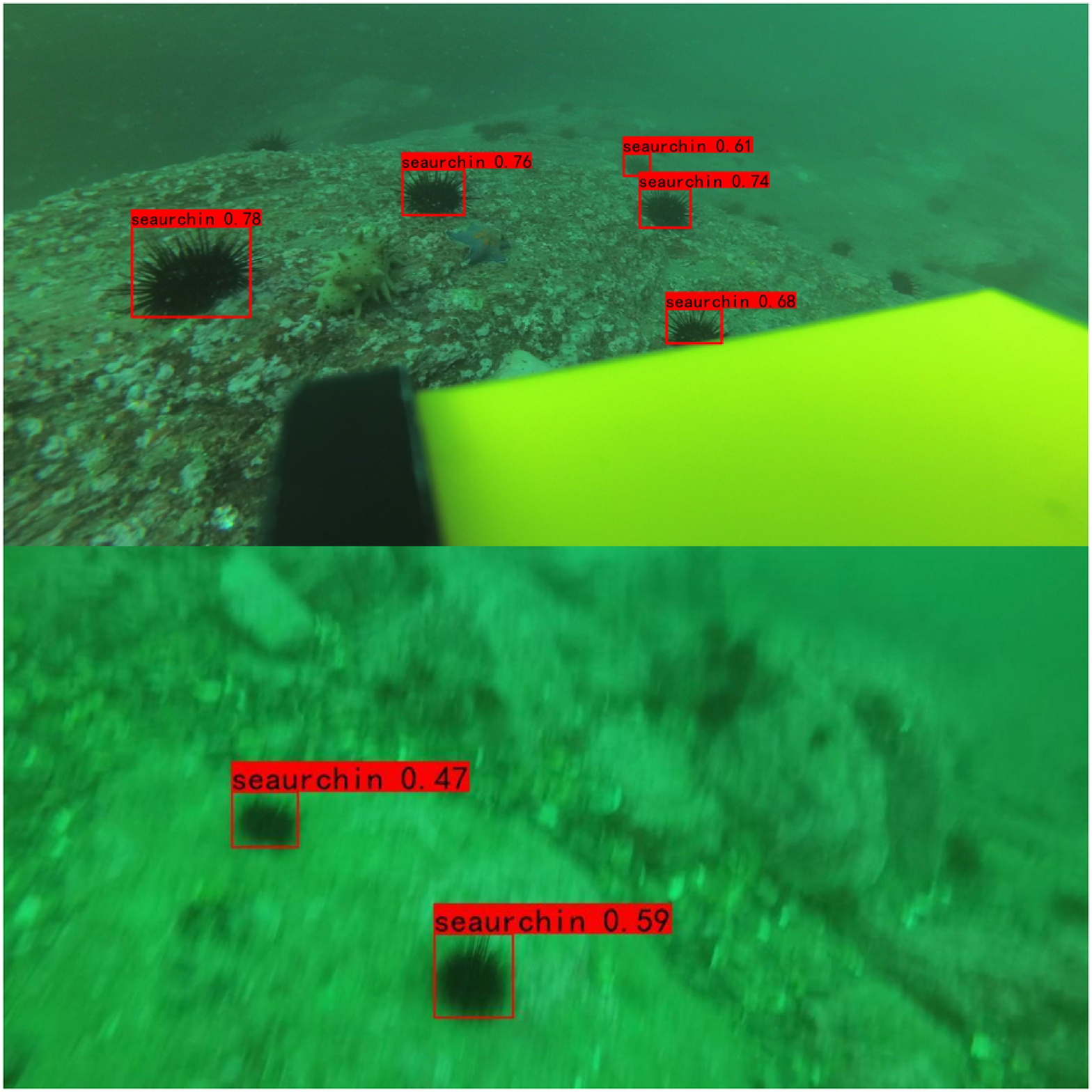}}
	\hspace{-0.6cm}
	\subfigure[\fontsize{5pt}{\baselineskip}\selectfont{FUnI-EGAN}]{
		\label{Fig. 9. sub. 2}
		\includegraphics[width=2cm,height=3cm]{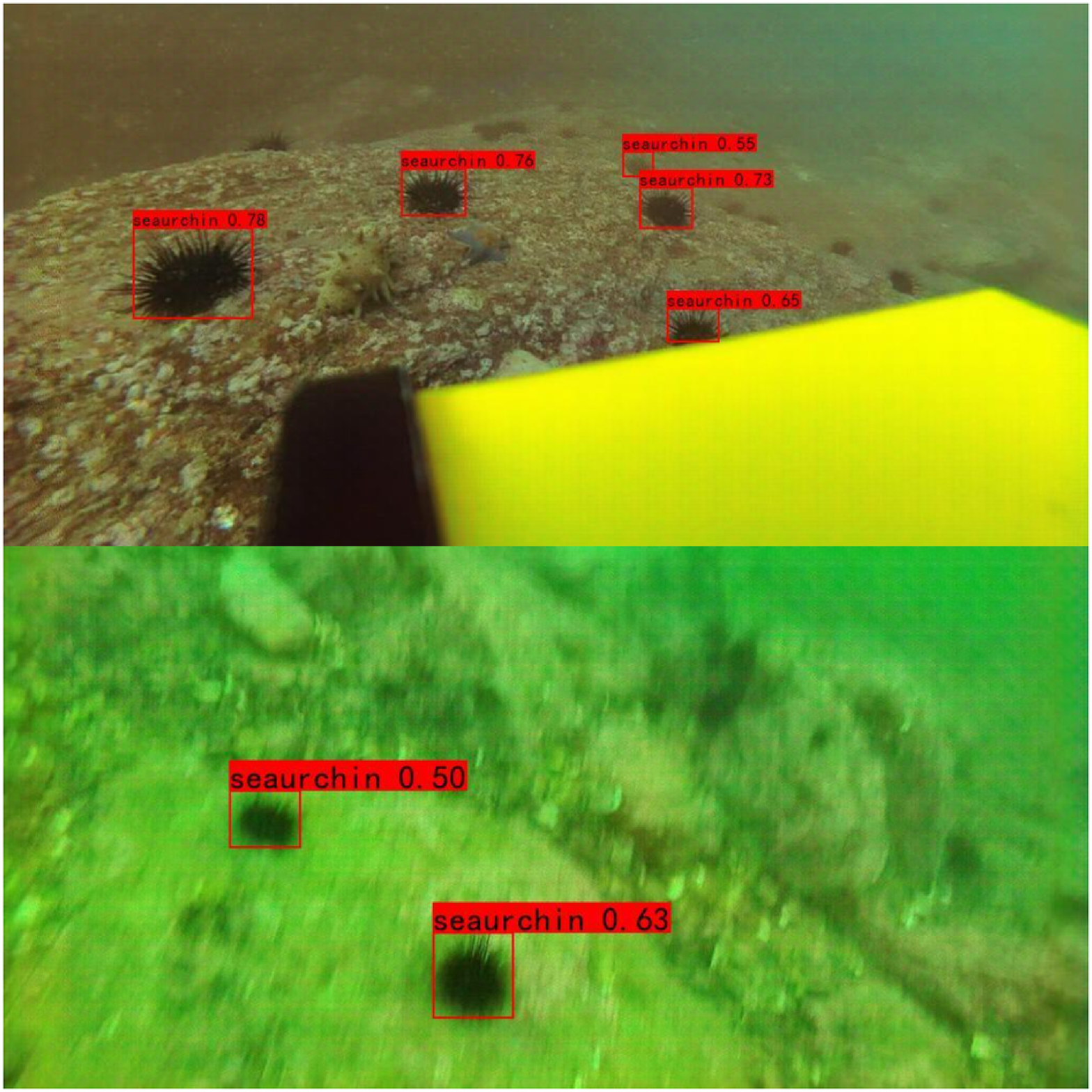}}
	\hspace{-0.6cm}
	\subfigure[\fontsize{7pt}{\baselineskip}\selectfont{SCNet}]{
		\label{Fig. 9. sub. 3}
		\includegraphics[width=2cm,height=3cm]{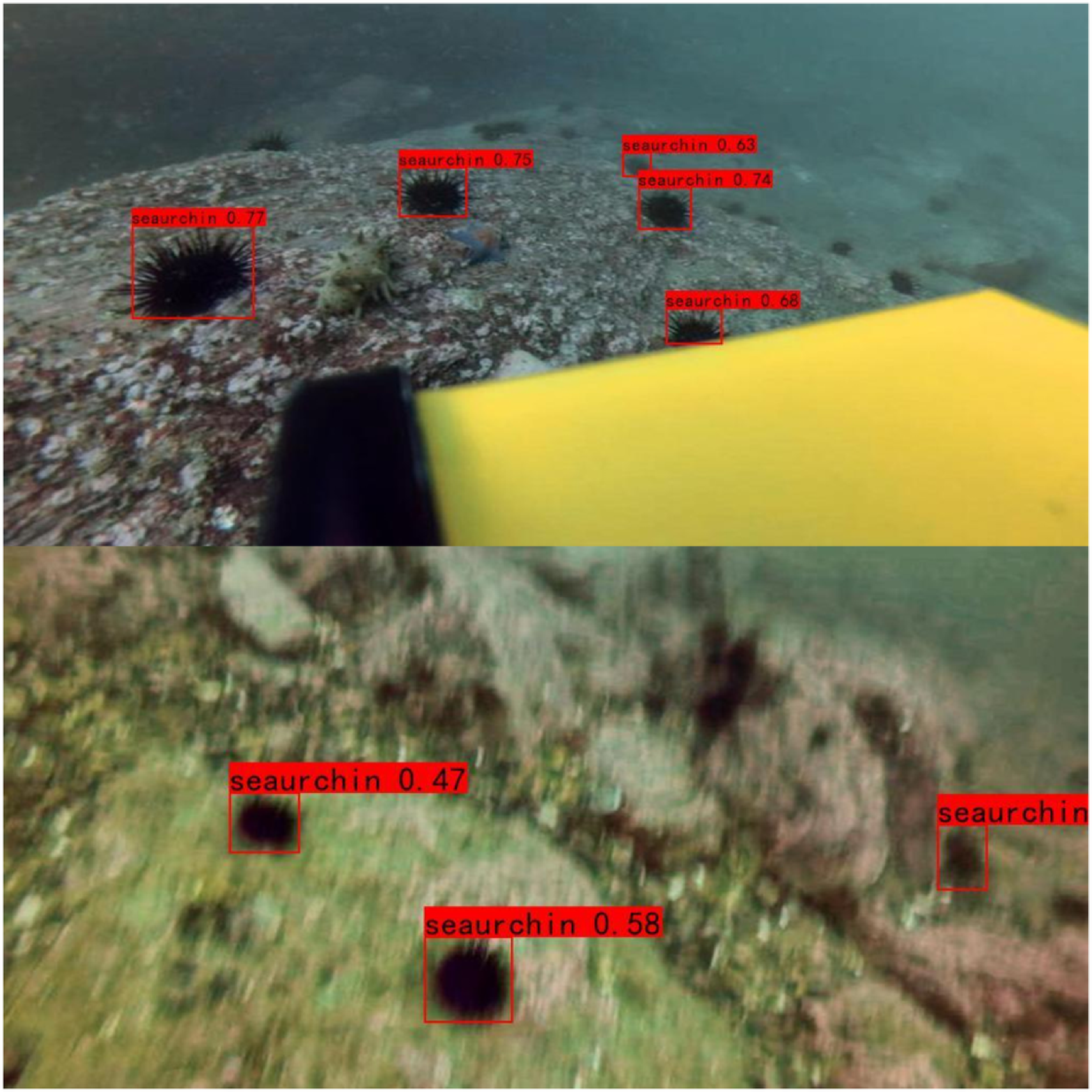}}
	\hspace{-0.6cm}
	\subfigure[\fontsize{7pt}{\baselineskip}\selectfont{CLAHE}]{
		\label{Fig. 9. sub. 4}
		\includegraphics[width=2cm,height=3cm]{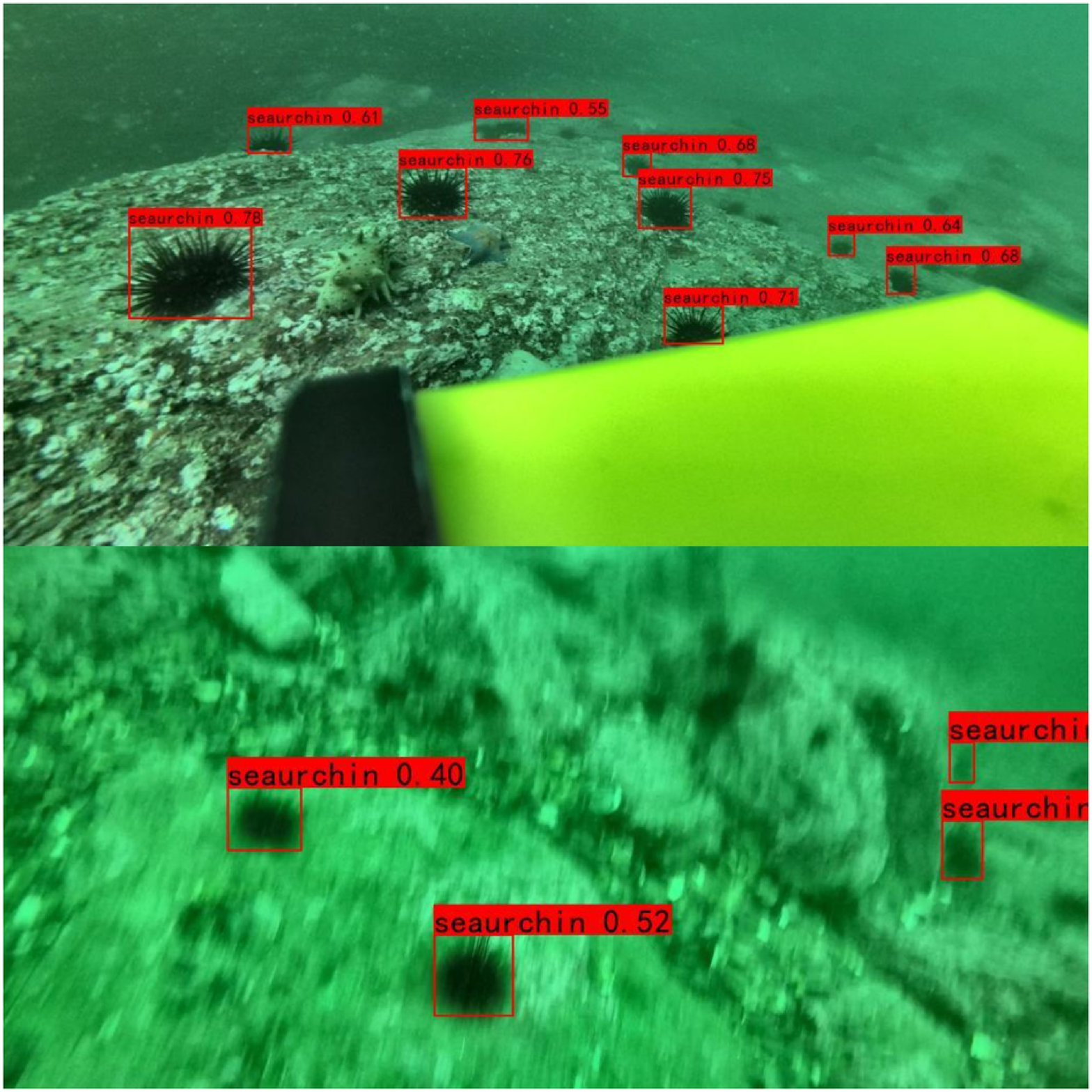}}
	\hspace{-0.6cm}
	\subfigure[\fontsize{7pt}{\baselineskip}\selectfont{MLLE}]{
		\label{Fig. 9. sub. 5}
		\includegraphics[width=2cm,height=3cm]{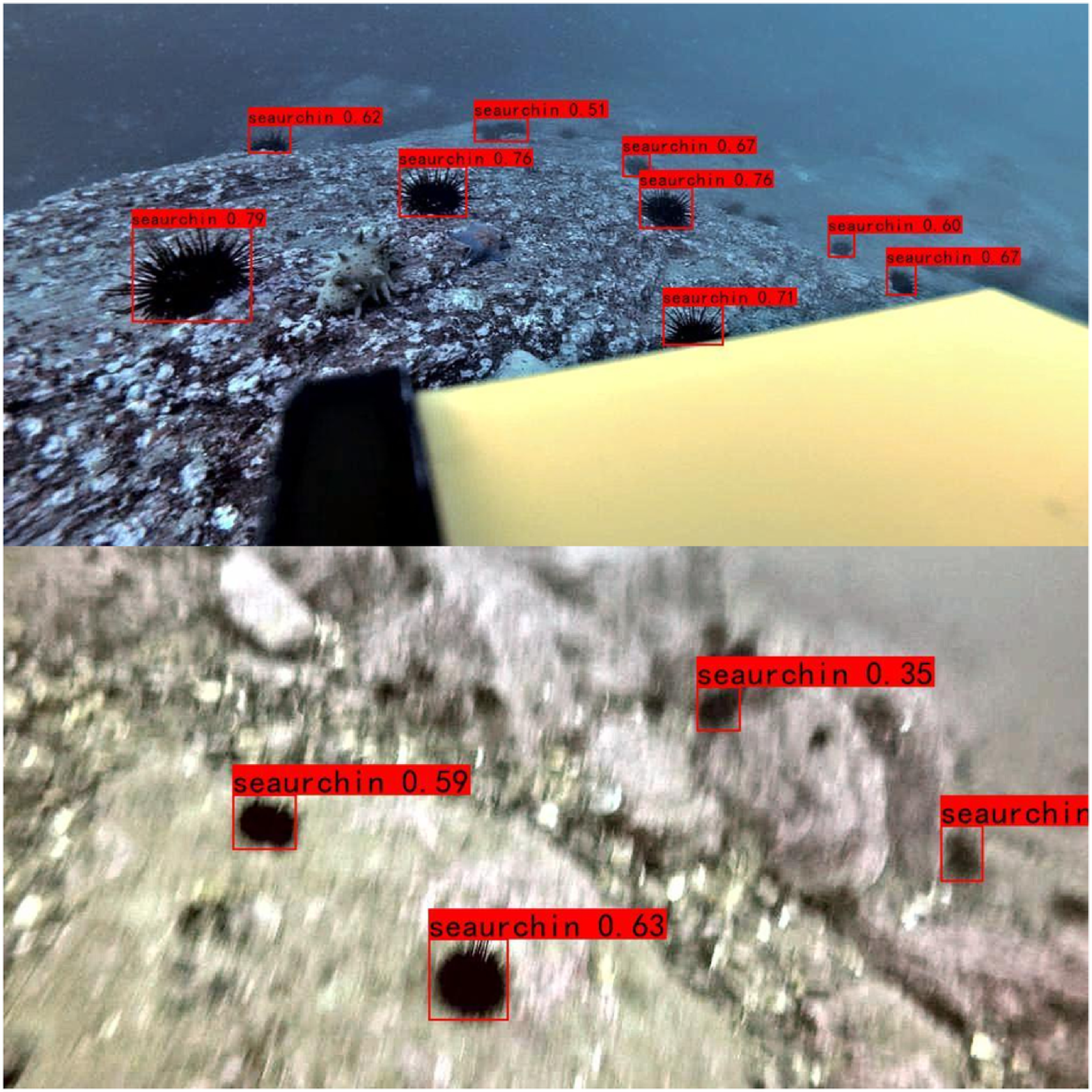}}
	\hspace{-0.6cm}
	\subfigure[\fontsize{7pt}{\baselineskip}\selectfont{UNVT}]{
		\label{Fig. 9. sub. 6}
		\includegraphics[width=2cm,height=3cm]{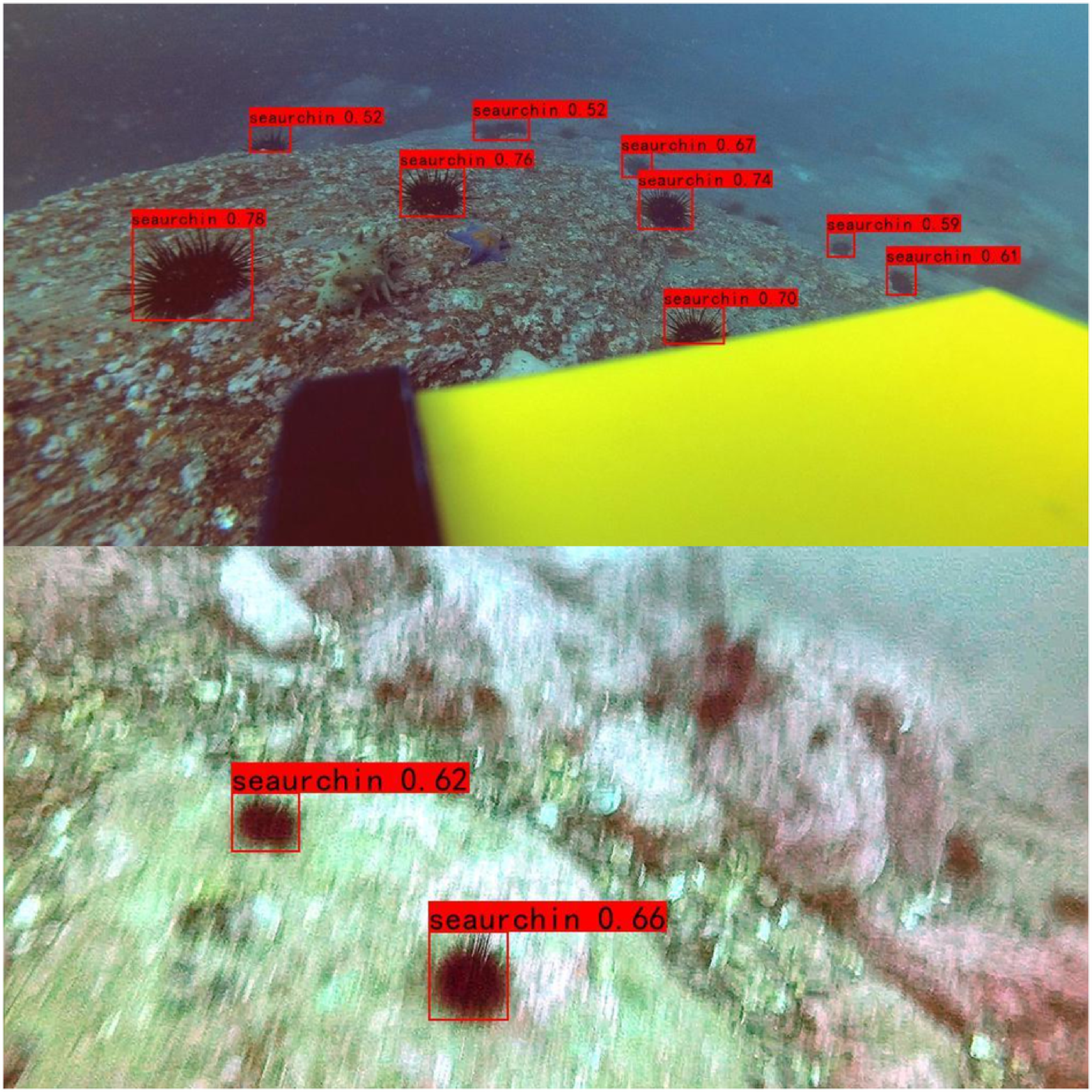}}
	\hspace{-0.6cm}
	\subfigure[\fontsize{7pt}{\baselineskip}\selectfont{Ours}]{
	\label{Fig. 9. sub. 7}
		\includegraphics[width=1.7cm,height=3cm]{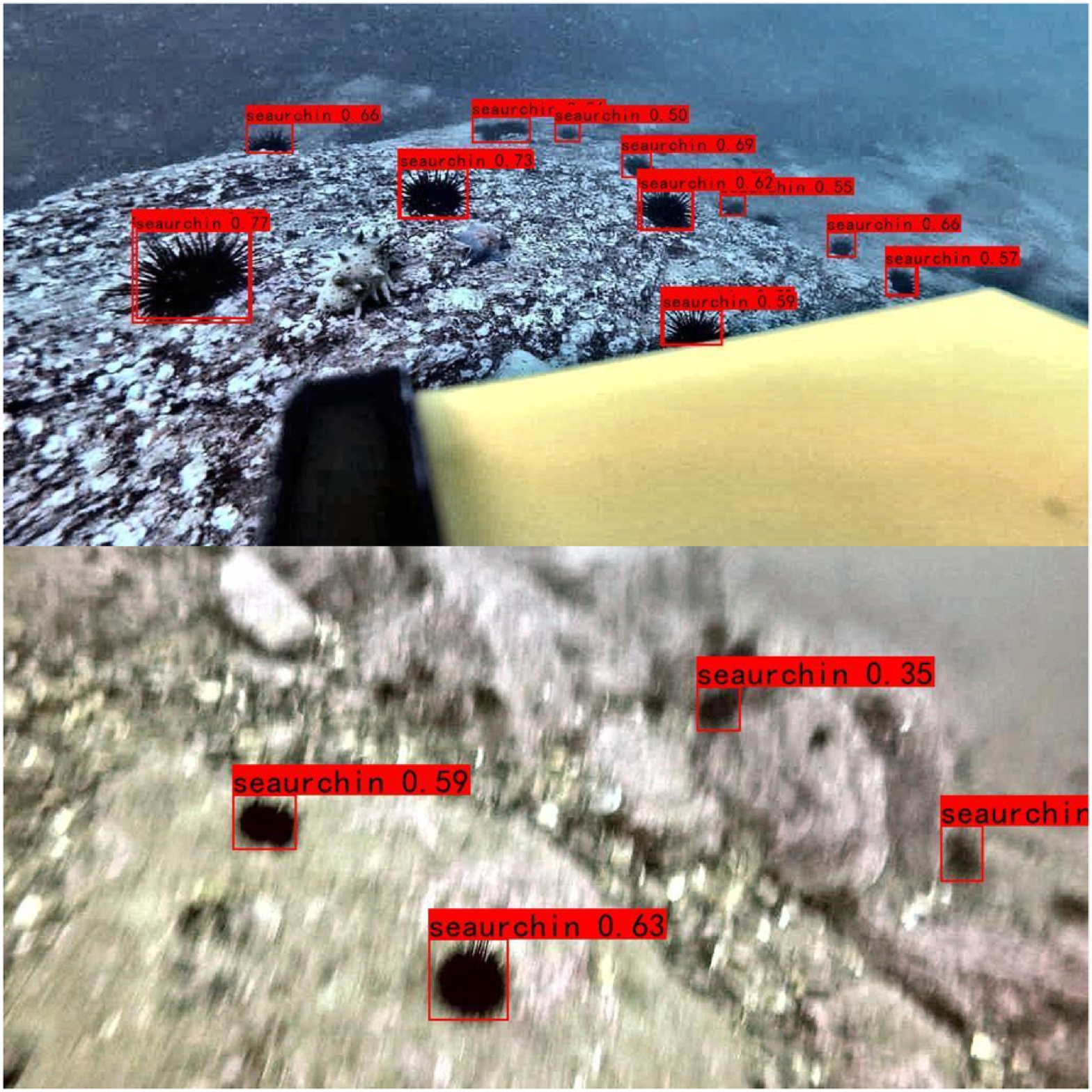}}
	\caption{Visual comparison of enhancement methods on UDD.}
	\label{Fig. 9}
\end{figure}

We compare our proposed method to the advanced underwater enhancement methods, including deep learning-based methods: FUnIE-GAN\cite{ref_article10}, SCNet\cite{ref_proc9} and physical model-based methods: CLAHE\cite{ref_proc8}, UNVT\cite{ref_article11}, and MLLE\cite{ref_article12} on UDD. Although these excellent underwater image enhancement methods are mainly designed to improve image visual quality, the researchers state that these methods can be beneficial to help high-level vision tasks or feature-matching tasks which is similar to our goal. The enhanced results are shown in Fig.~\ref{Fig. 9}.

Compared to physical model-based enhancement methods, deep learning-based enhancement methods generally have weaker generalization ability. The deep convolutional neural network performs better in color correction, but less well at contrast correction which is precisely contrary to image utility quality tendency. FUnIE-GAN, for example, enhances image visual quality while weakening image gradient features, leading to a decrease in image utility quality instead. Table~\ref{tab8} provides the performance comparison of each enhancement method.

\begin{table}[t]
	\caption{Performance comparison of enhancement methods on UDD}
	\begin{center}
		\renewcommand\arraystretch{1.6}
		\begin{tabular}{l|l|l|l|l|l|l|l}
			\hline
			Method & raw & FUnIE-GAN & SCNet & CLAHE& UNVT &MLLE & Ours \\ 
			\Xhline{1pt}
			 mAP &75.13 & 69.67 &80.41 & 86.36 & 83.42 & 87.38 & \textbf{90.99} \\  
			 UIQM &0.05 & 0.69 & 0.60 & 0.46& 1.08& 1.04&\textbf{1.09} \\  
			 UCIQE&21.47 & 26.16 & 25.40 & 24.63& 29.26& 29.43&\textbf{29.47}  \\  
			FPS &  & 25 & 30 & 37 & 5 &9& \textbf{105} \\ 	
			\hline
		\end{tabular}
		\label{tab8}
	\end{center}
\end{table}

To additionally quantitatively evaluate image visual quality, we introduce two no-reference evaluation metrics: UIQM and UCIQE. Experimental data illustrate that the improvement in visual quality does not coincide with the improvement in utility quality. Our proposed method outperforms advanced enhancement methods by 3.61$\% $. At the same time, the time complexity of our method makes it well-suited for underwater scenes.

\section{Conclusion}
In this paper, we propose an underwater image enhancement method, that selectively enhances useful image features, and improves image utility quality. The task-oriented dynamic enhancement framework can be widely applied before various detection tasks. In comparison to the advanced underwater image enhancement methods, our method shows promising results. Since image utility quality can offer recommendations and guidance for boosting high-level vision task performance. Our work opens up many possibilities for further exploration. Inspired by reinforcement learning, in future work, we plan to develop a feedback-optimization enhancement strategy that uses image utility quality as supervision.

\subsubsection{Acknowledgements} This work was supported in part by the National Natural Science Foundation of China under Grant 61901119, and in part by the Natural Science Foundation of Fujian Province under Grant 2022J05117.

\bibliographystyle{unsplncs04}
\bibliography{ref}

\end{document}